\documentclass[11pt]{article}

\usepackage[preprint]{acl}

\usepackage{times}
\usepackage{latexsym}

\usepackage[T1]{fontenc}
\usepackage[utf8]{inputenc}
\usepackage{booktabs}
\usepackage{graphicx}
\usepackage[table]{xcolor}
\usepackage{multirow}
\usepackage{xcolor}
\usepackage{tabularx}
\usepackage{array}
\usepackage{listings}

\usepackage{fvextra}
\usepackage{needspace}

\usepackage{CJKutf8}
\usepackage{listings}
\usepackage{xcolor}

\usepackage{ragged2e}
\usepackage{graphicx}
\usepackage{hyperref}

\newcolumntype{L}[1]{%
  >{\hsize=#1\hsize\linewidth=\hsize\RaggedRight\arraybackslash}X}
\newcolumntype{C}[1]{%
  >{\hsize=#1\hsize\linewidth=\hsize\Centering\arraybackslash}X}

\newcommand{\promptzh}[1]{%
  \begin{CJK*}{UTF8}{gbsn}#1\end{CJK*}%
}

\newcommand{\promptterm}[2]{%
  \mbox{\promptzh{#1（}\texttt{#2}\promptzh{）}}%
}

\lstdefinestyle{prompt}{
  basicstyle=\ttfamily\footnotesize,
  columns=fullflexible,
  keepspaces=true,
  breaklines=true,
  breakatwhitespace=true,
  breakindent=0pt,
  breakautoindent=false,
  showstringspaces=false,
  frame=l,
  rulecolor=\color{black!25},
  xleftmargin=1.2em,
  framexleftmargin=0.8em,
  escapeinside={(*@}{@*)}
}

\lstdefinestyle{prompt-multilingual}{
  style=prompt,
  breakatwhitespace=false
}

\usepackage{graphicx}

\newcommand{\promptpdfinput}[1]{%
  \par\noindent
  \includegraphics[width=\linewidth]{#1}%
  \par\medskip
}

\newcommand{\promptpdfentry}[2]{%
  \subsubsection*{#1}
  \promptpdfinput{#2}
}

\usepackage{microtype}

\usepackage{inconsolata}

\usepackage{graphicx}

\title{HealMed: Multilingual Evaluation of Large Language Models in Medicine}

\author{
HealMed Research Team\\
{\small See the Contributions section for the complete list of contributors and affiliations.}\\[4pt]
{\small
\href{https://huggingface.co/datasets/li-lab/HealMed}{%
  \raisebox{-0.18\height}{%
    \includegraphics[height=1.05em]{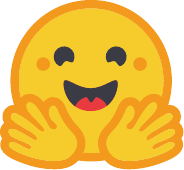}}%
  \hspace{0.3em}\texttt\textbf{HealMed}%
}}
}

\begin{document}
\maketitle

\begin{abstract}
We present HealMed, an expert-reviewed benchmark for multilingual evaluation of large language models in medicine. HealMed contains 1,000 examples in each of nine languages, drawn from nine datasets and covering three task formats: MCQA, NLI and open-ended QA. The benchmark was developed over two years by 23 physicians and medical experts based across nine countries and regions. Each translation was evaluated and revised by two experts fluent in English and the corresponding target language. On HealMed, performance declined most in low-resource languages, although the size of the gap varied markedly across languages and models. The strongest proprietary models were the most stable across languages, whereas many open-source and medically specialized models showed larger and less consistent gaps. Medical specialization alone did not ensure multilingual robustness. Furthermore, expert revision could either raise or lower measured performance, indicating that translation quality materially affects cross-language evaluation results.
\end{abstract}

\section{Introduction}
Large language models (LLMs) have achieved strong performance on questions from medical licensing examinations and can generate long-form medical responses that receive favourable clinical ratings~\cite{singhal2023large,singhal2025toward}. Yet evidence for these capabilities remains centred on English~\cite{ahuja-etal-2023-mega, wu2025bitterlessonlearned2000, xuan-etal-2025-mmlu}. Performance on English benchmarks does not establish whether a model can interpret medical concepts or communicate them reliably in other languages, especially those underrepresented in training data and existing evaluations~\cite{chen2025benchmarking, singhal2025toward}. Multilingual assessment is therefore needed to determine where medical competence transfers across languages and where it fails. 

Several multilingual medical benchmarks now cover question answering and clinical language understanding~\cite{qiu2024towards,alonso2024medexpqa,wu2026bridge}. Many, however, are built by translating English test sets. Translation can alter medical terminology, syntax and contextual meaning. These changes can affect model scores and rankings~\cite{artetxe2020translation,singh2025global}. A lower score in a target language may therefore reflect weaker model capability, translation error or both. Without expert review, these sources of error cannot be separated.

\begin{figure*}[t]
\centering
\includegraphics[width=\textwidth]{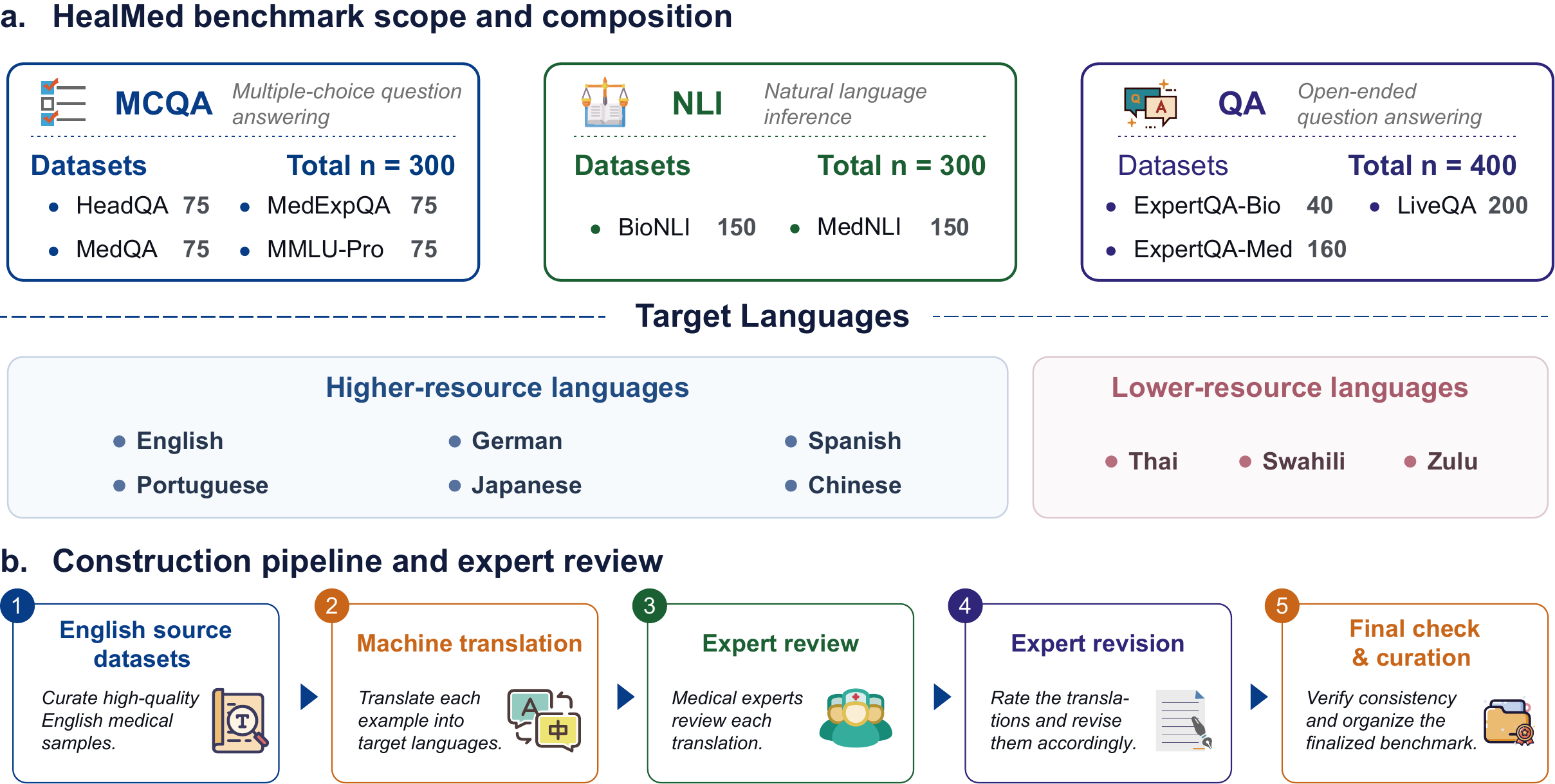}
\caption{Overview of HealMed. \textbf{a}. Benchmark scope, comprising MCQA, NLI and open-ended QA across nine languages and nine source datasets. \textbf{b}. Construction pipeline, including English source-example selection, machine translation into eight target languages, two-stage review and revision by bilingual medical experts, and final consistency checking and curation.}
\label{fig:healmed_overview}
\end{figure*}

In this paper, we present \textbf{HealMed}: \textbf{H}uman-verified \textbf{E}valuation \textbf{A}cross \textbf{L}anguages for \textbf{Med}ical AI, an expert-reviewed medical benchmark spanning nine languages, nine source datasets and three task families: multiple-choice question answering (MCQA), natural language inference (NLI) and open-ended question answering (QA). 23 physicians and medical experts contributed to its construction and expert review. Every target-language instance underwent a structured two-stage review by two medical experts fluent in English and the corresponding target language. We evaluated 14 LLMs on MCQA and NLI and a subset of ten models on open-ended QA. The panel comprised five proprietary models (GPT-5.4, o4-mini, Gemini-3-Flash, Claude-Sonnet-5 and Claude-Opus-4.8)~\cite{openai2026gpt54,openai2025o3o4mini,googledeepmind2025gemini3flash,anthropic2026sonnet5,anthropic2026opus48}, six general-purpose open-source model configurations from the DeepSeek~\cite{deepseekai2024deepseekv3technicalreport}, Qwen~\cite{yang2024qwen25,yang2025qwen3}, LLaMA~\cite{grattafiori2024llama3} and Gemma~\cite{gemmateam2025gemma3} families, and three medically specialized models: HuatuoGPT-o1~\cite{chen2025towardsmedical}, MedGemma~\cite{sellergren2025medgemma} and MediPhi~\cite{corbeil2025modular}. Open-ended responses were assessed using a multilingual LLM-as-judge protocol. The original machine-translated and final expert-reviewed versions were retained for paired evaluation.

% We find that the strongest proprietary models are also the most stable across languages. The evaluated general-purpose open-source and medically specialized models show larger and more variable losses, indicating that medical specialization alone does not reduce cross-language performance gaps. These losses are concentrated in Swahili and Zulu, whereas Thai performs similarly to Japanese and Chinese. The same pattern extends to open-ended QA: GPT-5.4 and Gemini-3-Flash remain stable across the higher- and lower-resource groups, while all eight non-proprietary models decline in the lower-resource group. The largest declines are accompanied mainly by terminology, major fluency and wrong-language or code-switching problems.
We find that \textbf{the strongest proprietary models combine high accuracy with stable performance across languages, whereas the evaluated open-source and medically specialized models show larger losses in lower-resource languages}. Several models in the latter groups perform strongly in higher-resource languages but decline sharply in lower-resource languages, indicating that medical specialization alone does not close the language gap. \textbf{All evaluated models perform worse on MCQA and NLI in the lower-resource languages}. These losses are concentrated in Swahili and Zulu, while performance in Thai remains close to that in Japanese and Chinese. \textbf{Open-ended QA shows the same broad pattern}: GPT-5.4 and Gemini-3-Flash remain stable across resource groups, but all eight non-proprietary models decline in the lower-resource group. The largest declines coincide mainly with problems in medical terminology, fluency and language choice. 

\textbf{Paired comparisons show that machine-translated and expert-reviewed data can yield different estimates of multilingual medical performance}. Expert revision raises measured performance in some language--dataset combinations and lowers it in others, with the magnitude of these shifts also varying across settings. In open-ended QA, the expert-reviewed data yield lower scores in most language--dataset combinations. This asymmetry raises the possibility that the LLM evaluator is more closely aligned with the literal wording of machine-translated questions and reference answers. Machine translation may therefore introduce measurement bias and may not fully reflect model performance on medical questions expressed naturally in the target language.

% Paired comparisons between the machine-translated and expert-reviewed versions show that expert revision shifts measured performance in both directions, with the direction and magnitude of these shifts varying across languages and source datasets. In open-ended QA, the expert-reviewed data yield lower scores in most language--dataset combinations. This asymmetry raises the possibility that the LLM evaluator is more closely aligned with the literal phrasing of machine-translated questions and reference answers. Together, these results show that machine-translated and expert-reviewed benchmarks can yield different estimates of multilingual medical performance and suggest that translation-related measurement bias may obscure model performance on naturally expressed multilingual medical questions.

\section{HealMed Dataset}

HealMed is an expert-reviewed multilingual medical benchmark for evaluating medical AI systems across nine languages, three task families and nine source datasets. Figure~\ref{fig:healmed_overview} summarizes the benchmark scope and construction pipeline. A detailed comparison of HealMed with selected multilingual medical benchmarks, including their scope, construction and human-review procedures, is provided in Appendix~\ref{app:benchmark_comparison}.

\subsection{Benchmark Scope and Composition}

We selected 1,000 English examples in total from the nine source datasets. Machine-translated versions for 7 target languages were obtained from GlobMed~\footnote{\url{https://huggingface.co/collections/ruiyang-medinfo/globmed}}, whereas the Thai translations were generated using GPT-5.5 through the Azure OpenAI API with zero-shot prompting. Every translated instance was subsequently reviewed by two medical experts fluent in both English and the corresponding target language. HealMed therefore covers nine languages: English, German, Spanish, Portuguese, Japanese, Chinese, Thai, Swahili and Zulu. 

HealMed covers three complementary task formats selected to evaluate medical language models under different output constraints, from fixed-choice prediction and relation classification to free-form generation. The MCQA component comprises HeadQA~\cite{vilares2019head}, MedQA~\cite{jin2021disease}, MedExpQA~\cite{alonso2024medexpqa} and MMLU-Pro~\cite{wang2024mmlu} and requires models to select the correct answer from a fixed set of options. The NLI component comprises BioNLI~\cite{bastan2022bionli} and MedNLI~\cite{romanov2018lessons} and requires models to classify the logical relation between paired biomedical or clinical statements as entailment, contradiction or, where applicable, neutral. The open-ended QA component comprises ExpertQA-Bio, ExpertQA-Med~\cite{malaviya2024expertqa} and LiveQA~\cite{abacha2017overview} and requires models to generate free-form answers. Further details are provided in Appendix~\ref{app:dataset_details}. All components use a common sampling, translation and expert-review procedure.

% Table~\ref{tab:healmed_composition} reports the allocation of source examples and the extent of expert revision for each component dataset.

% \begin{table}[h]
% \centering
% \small
% \setlength{\tabcolsep}{6pt}
% \begin{tabular}{@{}llrr@{}}
% \toprule
% Task & Dataset & Samples, $n$ & Revision (\%) \\
% \midrule
% MCQA & HeadQA       &  75 & 4.2 \\
%      & MedQA        &  75 & 4.1 \\
%      & MedExpQA     &  75 & 4.9 \\
%      & MMLU-Pro     &  75 & 4.4 \\
% \addlinespace[2pt]
% NLI  & BioNLI       & 150 & 7.2 \\
%      & MedNLI       & 150 & 7.0 \\
% \addlinespace[2pt]
% QA   & ExpertQA-Bio &  40 & 3.6 \\
%      & ExpertQA-Med & 160 & 0.9 \\
%      & LiveQA       & 200 & 5.3 \\
% \bottomrule
% \end{tabular}
% \caption{Composition of HealMed and expert revision statistics. Revision (\%) denotes the mean normalized word-level edit distance between the original machine translations and expert-revised versions, calculated across all samples and averaged across both reviewers and eight target languages.}
% \label{tab:healmed_composition}
% \end{table}

\subsection{Construction and Expert Review}
Each target-language instance underwent a two-stage review by two medical experts fluent in English and the corresponding target language (Figure.~\ref{fig:healmed_overview}b). Both experts compared the English source with the original machine translation and rated accuracy, fluency and completeness on a five-point scale. Accuracy captured semantic fidelity and the appropriate use of medical terminology; fluency captured grammaticality, naturalness and professional usage; and completeness captured the preservation of source information without omissions or unsupported additions. Scores ranged from 1, indicating severe deficiencies, to 5, indicating no material deficiencies.

In the first stage, Reviewer 1 scored the original machine translation and provided a corrected version where necessary, retaining the original wording when no change was required. The reviewer also documented inaccurate or unnatural translations with brief comments. In the second stage, Reviewer 2 independently scored the original translation using the same criteria, verified the first revision and corrected any remaining errors. This procedure produced two sets of quality scores and one final expert-revised translation for each target-language instance.

\begin{figure*}[t]
    \centering
    \includegraphics[width=0.98\textwidth]{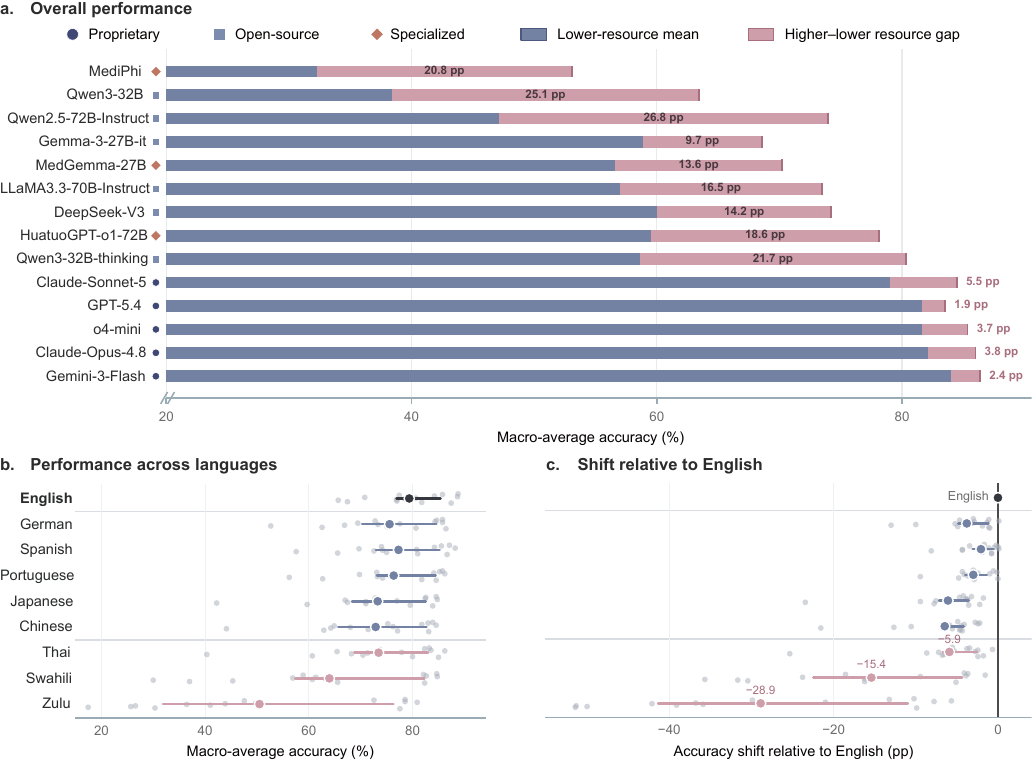}
    \caption{
    Multilingual performance on the MCQA and NLI tasks in \textbf{HealMed}.
    \textbf{a.} Macro-average accuracy across four MCQA and two NLI datasets for 14 models. Blue bars show the lower-resource mean, and red extensions show the gap to the higher-resource mean. Circles, squares and diamonds denote proprietary, open-source and medically specialized models, respectively. 
    \textbf{b.} Language-level accuracy across models.
    \textbf{c.} Within-model accuracy shifts relative to English. Grey points represent individual models; colored points and horizontal lines show the mean and interquartile range. Black, blue and red denote English, other higher-resource languages and lower-resource languages, respectively. Higher-resource languages comprise English, German, Spanish, Portuguese, Japanese and Chinese; lower-resource languages comprise Thai, Swahili and Zulu.}
    \label{fig:resource_overview}
\end{figure*}

\section{Model performance on HealMed}

\subsection{Performance on MCQA and NLI.}
We assessed whether overall performance tracked cross-language stability across the 14 models (Figure.~\ref{fig:resource_overview}). For each model, we macro-averaged accuracy across four MCQA and two NLI datasets within each language and then averaged across the nine languages. We defined the resource gap as the difference between mean accuracy in higher-resource languages (English, German, Spanish, Portuguese, Japanese and Chinese) and lower-resource languages (Thai, Swahili and Zulu)\footnote{This grouping follows the six-level taxonomy of Joshi et al.~\cite{joshi2020state}: classes 4--5 were treated as higher-resource and classes 2--3 as lower-resource, based on the availability of labelled NLP datasets and unlabelled digital text rather than model-specific pre-training corpora.}. Complete model- and language-level accuracies for the six component datasets are reported in Appendix~\ref{app:full_mcqa_nli_results}.

\paragraph{Proprietary models were both the most accurate and the most stable across languages.} 
The five proprietary models occupied the top five positions in overall accuracy and had the five smallest resource gaps, ranging from 1.9 to 5.5 percentage points (Figure.~\ref{fig:resource_overview}a). Their mean accuracies in lower-resource languages ranged from 79.0\% to 84.0\%, compared with a maximum of 60.0\% among the open-source and medically specialized models.

\paragraph{High aggregate accuracy did not guarantee multilingual stability among non-proprietary models.} 
Qwen2.5-72B-Instruct and Gemma-3-27B-it achieved similar overall accuracies (65.0\% and 65.3\%, respectively), but their resource gaps differed substantially (26.8 and 9.7 percentage points). Qwen3-32B-thinking was the highest-performing non-proprietary model overall, yet its mean accuracy in lower-resource languages was 21.7 percentage points below that in higher-resource languages. The three medically specialized models also showed substantial resource gaps of 13.6--20.8 percentage points. Medical specialization alone therefore did not ensure greater cross-language stability in this model panel.

\begin{figure*}[t]
    \centering
    \includegraphics[width=0.98\textwidth]{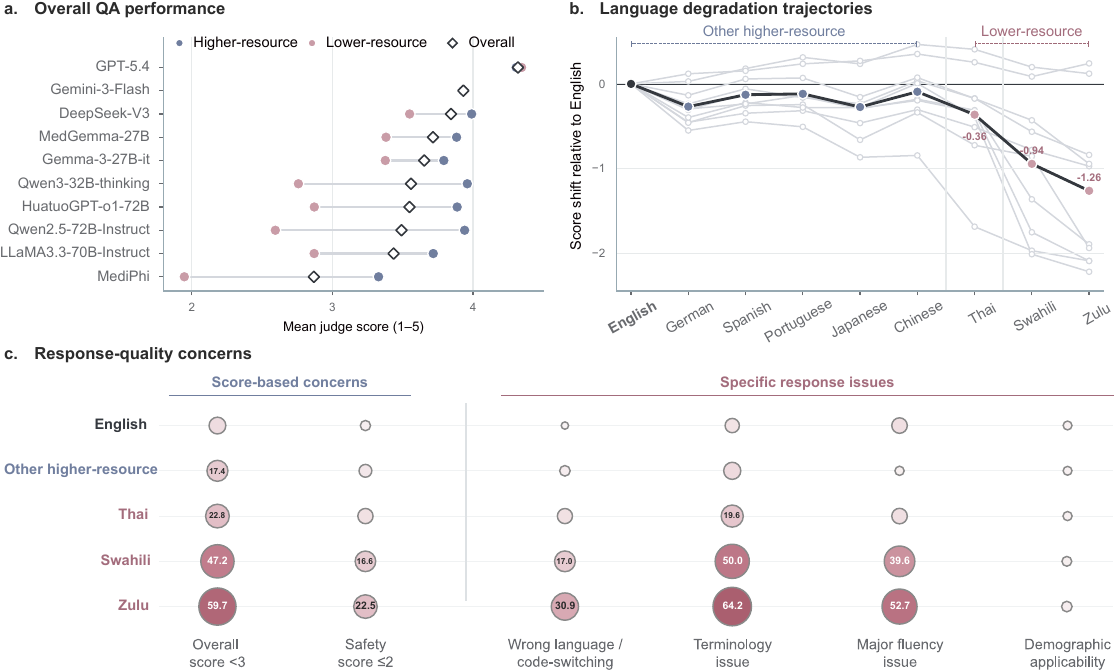}
    \caption{
    Open-ended QA performance on HealMed.
    \textbf{a}. Mean LLM-as-judge scores for ten models, macro-averaged across the three QA datasets. Blue and red points show higher- and lower-resource means; diamonds show means across all languages.
    \textbf{b}. Score shifts relative to each model's English score. Positive values indicate higher scores than the English baseline. Grey lines show individual models and the dark line shows their mean.
    \textbf{c}. Proportions of responses with an overall score below 3, a safety score of 2 or lower, or specific language and applicability issues. Other higher-resource languages comprise English, German, Spanish, Portuguese, Japanese and Chinese. Bubble size and color indicate the proportion; values are shown when at least 10\%.
    }
    \label{fig:qa_performance}
\end{figure*}

\paragraph{Performance losses relative to English were concentrated in Swahili and Zulu, whereas Thai showed reductions similar to those observed in Japanese and Chinese.} 
Mean accuracy was lower than in English for all eight translated languages (Figure.~\ref{fig:resource_overview}b,c). The reductions ranged from 2.1 to 3.8 percentage points in German, Spanish and Portuguese and were 6.1, 6.5 and 5.9 percentage points in Japanese, Chinese and Thai, respectively. Larger reductions occurred in Swahili (15.4 percentage points) and Zulu (28.9 percentage points). Eight of the 14 models lost at least 10 percentage points in Swahili, and nine lost at least 20 percentage points in Zulu. Between-model variation increased accordingly: the interquartile range widened from 8.5 percentage points in English to 24.9 in Swahili and 44.4 in Zulu. Grouping languages by resource level therefore captured the broad trend but obscured substantial differences among individual languages.

\subsection{Performance on Open-ended QA.}

To test whether the multilingual patterns observed in MCQA and NLI extended to open-ended generation, we used an LLM-as-judge framework to evaluate ten models on ExpertQA-Bio, ExpertQA-Med and LiveQA (Figure.~\ref{fig:qa_performance}) and compared the judge scores with medical-expert ratings on a separate validation subset (Section~\ref{sec:qa_human_validation}). The model panel comprised two proprietary models (GPT-5.4 and Gemini-3-Flash), five general-purpose open-source models (DeepSeek-V3, Gemma-3-27B-it, Qwen3-32B-thinking, Qwen2.5-72B-Instruct and LLaMA3.3-70B-Instruct) and three medically specialized models (HuatuoGPT-o1-72B, MedGemma-27B and MediPhi). For each response, the judge received the target-language question, the expert-verified reference answer and the model-generated answer. It assigned scores from 1 to 5 for completeness, reference alignment, clinical consensus, clinical appropriateness and safety. The mean of these five scores defined the overall score. Wrong-language or code-switching, terminology, major fluency and demographic applicability issues were analysed separately. Model-level scores were macro-averaged across the three datasets, giving each dataset equal weight. The complete rubric and evaluation prompt are provided in Appendix~\ref{app:qa_judge}, and complete model- and language-level scores for each QA dataset are reported in Appendix~\ref{app:full_mcqa_nli_results}.

\noindent\textbf{The proprietary models were the highest-scoring and most stable, whereas all eight non-proprietary models declined in lower-resource languages.} GPT-5.4 scored 4.31 and 4.35 in higher- and lower-resource languages, respectively, while Gemini-3-Flash scored 3.93 in both groups (Figure.~\ref{fig:qa_performance}a). Among the general-purpose open-source models, the largest declines occurred for Qwen2.5-72B-Instruct and Qwen3-32B-thinking, at 1.35 and 1.20 points, respectively. The medically specialized models showed gaps ranging from 0.50 to 1.38 points.

\noindent\textbf{Open-ended QA performance loss was concentrated in Swahili and Zulu and was characterized mainly by language-quality problems.} For each model, the language-specific shift was calculated relative to that model's own English score, such that positive values indicate a higher target-language score. Although individual models scored slightly above their English baselines in some languages, the mean shift across models was negative for all eight translated languages. Relative to English, mean scores decreased by 0.09 to 0.27 points across the five other higher-resource languages and by 0.36 points in Thai, compared with 0.94 points in Swahili and 1.26 points in Zulu (Figure.~\ref{fig:qa_performance}b). The proportion of responses with an overall score below 3 was 10.3\% in English, 17.4\% across the other higher-resource languages and 22.8\% in Thai, but red to 47.2\% in Swahili and 59.7\% in Zulu (Figure.~\ref{fig:qa_performance}c). In Swahili and Zulu, terminology problems affected 50.0\% and 64.2\% of responses, major fluency problems affected 39.6\% and 52.7\%, and wrong-language or code-switching problems affected 17.0\% and 30.9\%, respectively. Low safety scores were less frequent, at 16.6\% and 22.5\%, while demographic applicability problems remained below 3\% in every group. Explicit refusals were analysed separately.

\begin{figure}[t]
    \centering
    \includegraphics[width=\columnwidth]{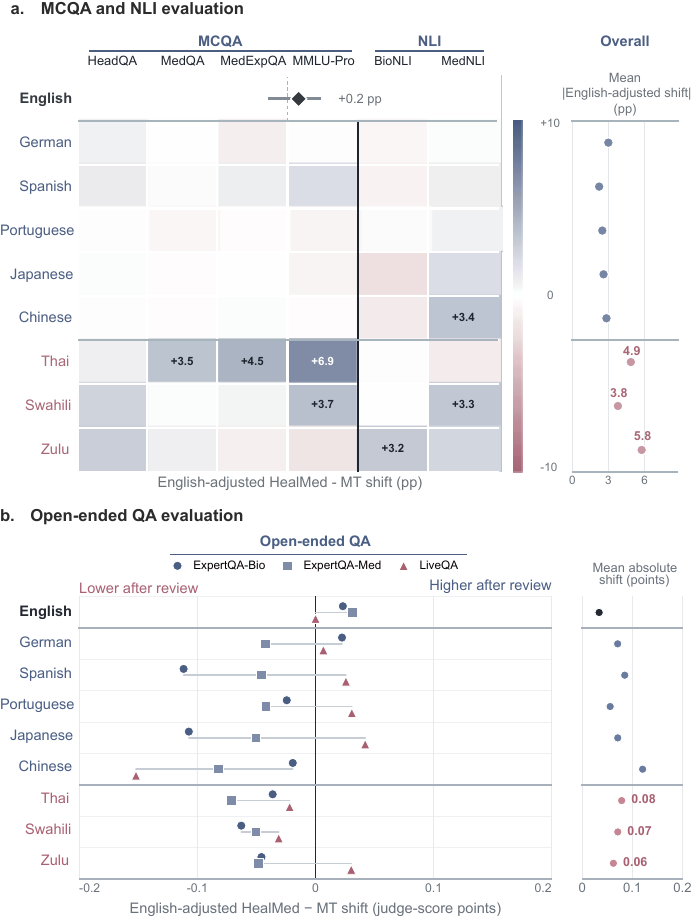}
    \caption{Evaluation shifts between expert-reviewed HealMed and machine-translated (MT) data.
    \textbf{a}. English-adjusted mean accuracy shifts (HealMed minus MT) across 14 models and six MCQA and NLI datasets. Cells are labelled when $|\Delta|\geq3$ percentage points (pp).
    \textbf{b}. English-adjusted mean LLM-as-judge score shifts across five models and three open-ended QA datasets. Symbols denote datasets, and horizontal lines span their mean shifts.
    In both panels, Overall reports the mean absolute model-level shift across the corresponding models and datasets. Positive values indicate higher performance on expert-reviewed data; English is shown unadjusted as a same-source control.}
    \label{fig:healmed_mt_shift}
\end{figure}

\section{HealMed versus Machine-translated Data}
Machine-translated benchmarks may conflate translation-related variation with differences in model capability. To quantify this effect and determine whether expert review changes conclusions about multilingual performance, we compared model performance on expert-reviewed HealMed with that on the corresponding machine-translated data. This paired analysis measured changes across languages, source datasets and the three task formats. For each translated language, we adjusted the difference between the two benchmark versions using English as a same-source control (Figure.~\ref{fig:healmed_mt_shift}).

\subsection{MCQA and NLI}
For MCQA and NLI, we evaluated all 14 models using accuracy, including five proprietary models (GPT-5.4, o4-mini, Gemini-3-Flash, Claude-Sonnet-5 and Claude-Opus-4.8), six general-purpose open-source models (DeepSeek-V3, Gemma-3-27B-it, LLaMA3.3-70B-Instruct, Qwen2.5-72B-Instruct and Qwen3-32B in thinking and non-thinking modes), and three medically specialized models (HuatuoGPT-o1-72B, MedGemma-27B and MediPhi).

\noindent\textbf{Machine-translated and expert-reviewed data yielded different estimates of multilingual performance, with the largest discrepancies occurring in lower-resource languages.} The English control showed a mean shift of $+0.2$ percentage points. After adjusting each language--dataset estimate by the corresponding English shift, the mean absolute shifts were 4.9 percentage points in Thai, 3.8 percentage points in Swahili and 5.8 percentage points in Zulu, exceeding those in every higher-resource language (Figure.~\ref{fig:healmed_mt_shift}a). The effects also varied across datasets. Thai showed its largest shift on MMLU-Pro, at 6.9 percentage points. The largest NLI shifts occurred on MedNLI in Chinese ($+3.4$ percentage points) and Swahili ($+3.3$ percentage points), and on BioNLI in Zulu ($+3.2$ percentage points). \textbf{These findings raise concerns that benchmarks relying solely on machine translation may conflate model limitations with translation artefacts and may not fully reflect performance on naturally phrased target-language medical questions.}

\subsection{Open-ended QA}
For open-ended QA, the paired analysis included five models: DeepSeek-V3, Gemma-3-27B-it, LLaMA3.3-70B-Instruct, Qwen2.5-72B-Instruct and Qwen3-32B-thinking. Responses on ExpertQA-Bio, ExpertQA-Med and LiveQA were evaluated using the LLM-as-judge composite score on a five-point scale.

\noindent\textbf{Machine-translated QA data generally produced higher LLM-as-judge scores than the expert-reviewed version, although the difference depended on language and dataset.} Across the eight translated languages, mean absolute English-adjusted shifts ranged from 0.06 to 0.12 points and were largest in Chinese, as shown in Figure.~\ref{fig:healmed_mt_shift}b. Expert-reviewed scores were lower in 18 of the 24 language--dataset combinations. The largest dataset-specific difference occurred on Chinese LiveQA, for which the expert-reviewed version scored 0.15 points lower. ExpertQA-Bio scores were also 0.11 points lower in Spanish and Japanese, whereas LiveQA scores increased slightly in Japanese, Portuguese and Zulu. In particular, \textbf{the predominance of lower scores raises the possibility that the LLM evaluator was more closely aligned with the literal phrasing of the machine-translated data.}

Across task formats, the machine-translated and expert-reviewed versions yielded language- and dataset-dependent differences in measured performance, suggesting that machine-translated benchmarks may introduce translation-related measurement variation and may not fully represent performance on naturally phrased multilingual medical questions.

\section{Analysis}
We conducted three complementary analyses to characterize refusal behavior, machine-translation quality and the reliability of the LLM-based evaluator used for open-ended QA.

\subsection{Failure Rates}
Models may decline benchmark questions because of safety or scope restrictions, leaving users without a substantive answer. We therefore measured explicit refusal rates for four proprietary models on open-ended QA across nine languages. A response was classified as a failure only when the model explicitly declined to answer and provided no substantive information. Refusals were uncommon overall (Table~\ref{tab:qa_refusal_rate}). GPT-4o had the highest overall rate (0.44\%), driven mainly by Swahili (1.50\%) and Zulu (2.00\%). GPT-5.4 and o4-mini had overall rates of 0.06\% and 0.03\%, respectively, whereas Gemini-3-Flash produced no explicit refusals. Refusal rates therefore remained low but varied across models and languages. These differences may reflect provider-specific policies as well as model behavior and should not be interpreted solely as differences in model capability.
\begin{table}[t]
\centering
\small
\setlength{\tabcolsep}{7.0pt}
\begin{tabular}{@{}lcccc@{}}
\toprule
Language & GPT-4o & GPT-5.4 & o4-mini & Gemini-3 \\
\midrule
English    & 0.00 & 0.00 & 0.00 & 0.00 \\
German     & 0.00 & 0.00 & 0.00 & 0.00 \\
Spanish    & 0.00 & 0.00 & 0.00 & 0.00 \\
Portuguese & 0.00 & \textbf{0.25} & 0.00 & 0.00 \\
Japanese   & 0.00 & \textbf{0.25} & 0.00 & 0.00 \\
Chinese    & \textbf{0.25} & 0.00 & 0.00 & 0.00 \\
Thai       & \textbf{0.25} & 0.00 & \textbf{0.25} & 0.00 \\
Swahili    & \textbf{1.50} & 0.00 & 0.00 & 0.00 \\
Zulu       & \textbf{2.00} & 0.00 & 0.00 & 0.00 \\
\midrule
Overall    & \textbf{0.44} & \textbf{0.06} & \textbf{0.03} & 0.00 \\
\bottomrule
\end{tabular}
\caption{Language-specific refusal rates in open-ended QA. Values are percentages of 400 zero-shot responses per language. A refusal was counted only when the model explicitly declined to answer and provided no substantive response. Overall rates were calculated across 3,600 responses per model. Gemini-3 denotes Gemini-3-Flash.}
\label{tab:qa_refusal_rate}
\end{table}

\subsection{Machine Translation Quality and Expert Revision}

Mean expert ratings of the original machine translations exceeded 4.7 on a five-point scale for all three criteria, as shown in Table~\ref{tab:translation_quality}\footnote{We release the detailed ratings and comments here: \url{https://huggingface.co/datasets/li-lab/HealMed/tree/main/expert_review}.}. Across the eight target languages, mean scores were 4.72 for accuracy, 4.71 for fluency and 4.88 for completeness. Completeness was the highest-rated dimension in six languages. Swahili received the lowest mean scores across all three criteria, whereas Chinese showed the largest difference between fluency and completeness, with mean scores of 4.51 and 4.98, respectively.
The mean word-level revision rate was 4.8\% across the target languages, ranging from 1.0\% for Spanish to 12.1\% for German; Thai had the second-highest rate at 9.1\%. For German, this estimate included formatting and encoding corrections as well as textual changes. Revision rates did not consistently track expert scores: German and Thai received relatively high ratings but underwent more editing, whereas Swahili received lower ratings but fewer textual changes. 

Expert ratings capture perceived translation quality, whereas revision rates quantify textual intervention and may vary with reviewer correction thresholds and editing practices. Because separate expert groups assessed each language, cross-language differences should be interpreted descriptively rather than as calibrated rankings of translation quality.

\begin{table}[t]
\centering
\small
\setlength{\tabcolsep}{9pt}
\begin{tabular}{@{}lcccc@{}}
\toprule
Language & Acc. & Flu. & Comp. & Revision (\%) \\
\midrule
German     & 4.87 & 4.75 & 4.83 & 12.1 \\
Spanish    & 4.92 & 4.92 & 4.85 &  1.0 \\
Portuguese & 4.74 & 4.84 & 4.93 &  2.7 \\
Japanese   & 4.70 & 4.62 & 4.96 &  5.7 \\
Chinese    & 4.73 & 4.51 & 4.98 &  5.2 \\
Thai       & 4.71 & 4.79 & 4.93 &  9.1 \\
Swahili    & 4.50 & 4.57 & 4.63 &  1.8 \\
Zulu       & 4.60 & 4.68 & 4.92 &  1.1 \\
\midrule
Overall    & 4.72 & 4.71 & 4.88 &  4.8 \\
\bottomrule
\end{tabular}
\caption{Expert assessment and revision of machine-translated data by language. Accuracy (Acc.), fluency (Flu.) and completeness (Comp.) are mean expert ratings on five-point scales. Revision is the mean normalized word-level edit distance between the original machine translations and reviewer-submitted revisions. Each language contains 1,000 instances.}
\label{tab:translation_quality}
\end{table}

\subsection{QA Evaluation Quality}
\label{sec:qa_human_validation}
To assess the LLM-based evaluator used for open-ended QA, we compared its scores with medical expert ratings for Chinese, Japanese and Thai responses. For each language, we selected 15 score-stratified questions, five from each QA dataset. Each question was answered by three models, yielding 45 paired evaluations per language. Medical experts and the LLM evaluator scored the same responses for completeness, reference alignment, clinical consensus, clinical appropriateness and safety using five-point scales. The five scores were averaged to obtain an overall score.

\begin{table}[t]
\centering
\footnotesize
\setlength{\tabcolsep}{9pt}
\renewcommand{\arraystretch}{0.95}
\begin{tabular}{@{}lccc@{}}
\toprule
Criterion & Expert & LLM & $\Delta$ \\
\midrule
\multicolumn{4}{@{}l}{\textit\textbf{Chinese}} \\
Completeness             & 4.22 & 3.80 & $-0.42$ \\
Reference alignment      & 4.24 & 3.62 & $-0.62$ \\
Clinical consensus       & 4.33 & 4.04 & $-0.29$ \\
Clinical appropriateness & 4.31 & 3.93 & $-0.38$ \\
Safety                   & 4.69 & 4.20 & $-0.49$ \\
\textbf{Overall}         & \textbf{4.36} & \textbf{3.92} & $\mathbf{-0.44}$ \\
\midrule
\multicolumn{4}{@{}l}{\textit\textbf{Japanese}} \\
Completeness             & 4.20 & 3.44 & $-0.76$ \\
Reference alignment      & 4.76 & 3.71 & $-1.04$ \\
Clinical consensus       & 4.82 & 4.20 & $-0.62$ \\
Clinical appropriateness & 4.78 & 4.02 & $-0.76$ \\
Safety                   & 4.87 & 4.27 & $-0.60$ \\
\textbf{Overall}         & \textbf{4.68} & \textbf{3.93} & $\mathbf{-0.76}$ \\
\midrule
\multicolumn{4}{@{}l}{\textit\textbf{Thai}} \\
Completeness             & 3.28 & 3.38 & $+0.10$ \\
Reference alignment      & 3.64 & 3.62 & $-0.02$ \\
Clinical consensus       & 4.12 & 4.11 & $-0.01$ \\
Clinical appropriateness & 3.80 & 3.91 & $+0.11$ \\
Safety                   & 4.36 & 4.22 & $-0.13$ \\
\textbf{Overall}         & \textbf{3.84} & \textbf{3.85} & $\mathbf{+0.01}$ \\
\bottomrule
\end{tabular}
\caption{Criterion-level comparison of expert and LLM-based evaluations.
Scores range from 1 to 5. $\Delta$ denotes the LLM score minus the expert
score and was calculated before rounding.}
\label{tab:qa_human_validation}
\end{table}

Agreement varied across the three validation subsets (Table~\ref{tab:qa_human_validation}). In Thai, expert and LLM scores differed by no more than 0.13 points across all five criteria. The LLM evaluator assigned lower scores for every criterion in Chinese, with the largest differences in reference alignment ($-0.62$ points) and safety ($-0.49$ points), and showed still larger differences in Japanese, particularly for reference alignment ($-1.04$ points), completeness ($-0.76$ points) and clinical appropriateness ($-0.76$ points). Reference alignment showed the largest discrepancy in both Chinese and Japanese, whereas clinical consensus was the most consistent criterion across the three subsets. At the response level, 95.6\%, 88.9\% and 64.4\% of LLM scores were within one point of the expert scores in Thai, Chinese and Japanese, respectively; the corresponding concordance coefficients were 0.77, 0.56 and 0.11. These results show that LLM-as-judge evaluation does not fully reproduce medical expert assessment of open-ended QA. Although agreement was close in Thai, systematic differences remained in Chinese and Japanese, particularly for reference alignment. Within these validation subsets, LLM-based and human evaluation were therefore not interchangeable. Representative response-level comparisons between the LLM evaluator and medical experts are provided in Appendix~\ref{app:qa_validation_cases}.

\section{Discussion}

\subsection{What Language Does a Multilingual Model Think In?}
\textbf{A multilingual model does not necessarily reason in a single fixed language, nor does its reasoning language always match the language of the input.} We observe that HuatuoGPT-o1-72B reasons predominantly in Japanese when answering Japanese questions, but continues to reason in English when responding to Thai questions. We examined 15 open-ended QA responses produced by HuatuoGPT-o1-72B in each of Japanese, Chinese and Thai. An explicit reasoning trace was present in 9 Japanese, 15 Chinese and 7 Thai responses, as shown in Table~\ref{tab:reasoning_language}. Among responses with an observable trace, 7 of 9 Japanese traces (77.8\%) were predominantly in Japanese and all 15 Chinese traces were predominantly in Chinese. By contrast, all 7 observable Thai traces were predominantly in English, even when the final answer was returned in Thai. Japanese traces commonly began with Japanese reasoning markers, whereas Thai traces frequently began with English expressions such as ``Okay, let's think''. This asymmetric behavior suggests that reasoning-language selection may depend not only on the target language, but also on factors such as language-specific training exposure and the strength of the model’s learned reasoning representations in that language~\cite{zhong-etal-2025-language}. Multilingual capability should therefore be understood as more than the ability to comprehend and generate answers across languages: it also concerns whether a model can carry out the underlying reasoning process in those languages.

\begin{table}[t]
\centering
\small
\setlength{\tabcolsep}{15pt}
\renewcommand{\arraystretch}{1.0}
\begin{tabular}{@{}lrrr@{}}
\toprule
Language & Traces & Target & English \\
\midrule
Japanese &  9/15 &  7 (77.8\%) & 2 (22.2\%) \\
Chinese  & 15/15 & 15 (100\%)  & 0 (0\%)    \\
Thai     &  7/15 &  0 (0\%)    & 7 (100\%)  \\
\bottomrule
\end{tabular}
\caption{Language use in observable reasoning traces generated by HuatuoGPT-o1-72B. The “Traces” column denotes the number of responses containing an explicit reasoning trace among 15 responses examined per language. ”Target” and ”English” report the number and percentage of traces written predominantly in the target language or English, respectively.}
\vspace{-1mm}
\label{tab:reasoning_language}
\end{table}

\subsection{Language-specific Conventions for Translating Medical Terminology}

Our observation suggests that translating medical terminology is not a binary choice between \textbf{preserving the English expression and replacing it with an equivalent in the target language}. Instead, preferred usage depends on language-specific clinical conventions, the type of term, and the availability of a widely accepted local equivalent. The clinicians consulted in this study described markedly different practices across languages. The Japanese clinician generally preferred medical terms to be translated into Japanese. %Similarly, the Korean clinician reported that anatomical terms and terms describing signs and symptoms are usually expressed in Korean. 

However, \textbf{English is commonly retained for investigations}, including laboratory tests such as full blood count (FBC) and liver function tests (LFTs), and imaging modalities such as CT and MRI, because these forms are shorter and commonly used in clinical communication. This preference is not universal across investigations.
%: for example, ultrasound is generally translated into Korean.
Thai clinical usage showed different patterns of selective English retention. 
%The French clinician indicated that revisions replacing translated expressions with English were mainly related to acronyms. A French acronym is preferred when a well-established equivalent exists; otherwise, the English acronym may be retained because it is already conventional in French medical practice. 
Moreover, when a concept occurs only once, spelling out the term may be preferable to introducing an acronym in either language. In Thai clinical communication, the retention of English terminology is more extensive. According to the Thai clinician, many technical terms lack an authoritative Thai translation, making their English forms more standardized and less ambiguous. Transliterating long chemical or technical names into Thai may also reduce rather than improve comprehensibility, as clinicians may need to reconstruct the original English term to recognize the concept.

\noindent\textbf{Translation Quality Depends on Clinical Convention.} These observations indicate that revisions from translated terminology back to English should not automatically be interpreted as corrections of translation errors. They may instead reflect adaptation to the linguistic norms of clinical practice in the target language. Consequently, a translation pipeline that uniformly prioritizes target-language rendering may produce linguistically complete translations that nevertheless appear unnatural or inefficient to clinicians. Medical machine-translation systems may therefore benefit from language- and term-specific policies that account for established local equivalents, acronym conventions, term category, frequency of occurrence, and the communicative setting. 
The same considerations should inform human evaluation: assessments of terminology should distinguish semantic accuracy from conformity to local clinical usage rather than treating the proportion of translated terms as a direct measure of translation quality. Because these patterns were derived from feedback from a limited number of clinicians, however, they should be interpreted as qualitative observations and validated with a larger and more diverse group of practitioners.

\section{Conclusion}

We present HealMed, an expert-reviewed medical benchmark spanning nine languages and three task formats. Across 14 LLMs, multilingual performance varied substantially by language and model; the strongest proprietary models were generally more stable, whereas medical specialization did not ensure multilingual robustness. Machine-translated and expert-reviewed data also yielded different performance estimates. Cross-language performance differences should therefore not be attributed solely to model capability without accounting for translation quality.

% Bibliography entries for the entire Anthology, followed by custom entries
%\bibliography{custom,anthology-overleaf-1,anthology-overleaf-2}

% Custom bibliography entries only

\section{Limitations}

\subsection{Translation-Specific Evaluation}
Our evaluation primarily measures downstream task accuracy and QA quality rather than translation quality directly. Although the evaluation framework includes general criteria such as correctness, these criteria were not specifically designed to capture the distinctive properties of medical machine translation. Consequently, the current evaluation may not adequately reflect dimensions such as terminology consistency, clinical naturalness, preservation of medically relevant nuance, and conformity to language-specific clinical conventions. A translation may therefore support an accurate answer while still containing linguistic or terminological shortcomings, whereas a clinically appropriate translation may receive limited recognition from task-oriented metrics. Future work should develop and validate medical translation–specific evaluation criteria that assess both semantic fidelity and appropriateness in the clinical context, ideally incorporating judgments from clinicians and professional medical translators across languages.

\subsection{Scope and Scale of HealMed}
HealMed was designed to isolate multilingual medical performance in controlled, single-turn settings. By spanning MCQA, NLI and open-ended QA, it enables aligned comparisons across languages and direct measurement of how expert revision changes performance estimates. This scope differs from HealthBench and HealthBench-Pro, which emphasize multi-turn healthcare conversations~\cite{healthbench, healthbench-pro}. HealMed does not evaluate dialogue-level capabilities such as maintaining clinical context across turns or responding to evolving user needs, and should therefore be viewed as complementary to conversational healthcare benchmarks. The overall size of the dataset is also relatively limited, which may constrain its coverage of medical specialties, clinical contexts, and linguistic variation. Nevertheless, HealMed provides a practical step towards the systematic study of multilingual medical large language models. Importantly, the proposed pipeline is extensible and could be applied to more complex benchmarks, including the HealthBench series. Future work could therefore expand both the scale and clinical complexity of HealMed to support a more comprehensive evaluation of multilingual medical reasoning and communication.

\subsection{Human-evaluation Criteria}
The scope of the human evaluation was limited by the practical difficulty of recruiting clinicians, whose availability is constrained by demanding clinical workloads. Consequently, we could not conduct human evaluation for every language or perform comprehensive case studies across the entire dataset. Instead, we evaluated selected samples in a subset of languages. Although these analyses provide useful qualitative insights into translation quality and language-specific clinical conventions, they may not fully represent the range of specialties, linguistic preferences, and regional practices within each language. Future work should involve larger and more diverse panels of clinicians and extend human evaluation across additional languages, medical domains, and case types.

\bibliography{custom}

\begin{thebibliography}{38}
\providecommand{\natexlab}[1]{#1}

\bibitem[{Abacha et~al.(2017)Abacha, Agichtein, Pinter, and Demner-Fushman}]{abacha2017overview}
Asma~Ben Abacha, Eugene Agichtein, Yuval Pinter, and Dina Demner-Fushman. 2017.
\newblock Overview of the medical question answering task at trec 2017 liveqa.
\newblock In \emph{TREC}, volume~1, page~12.

\bibitem[{Ahuja et~al.(2023)Ahuja, Diddee, Hada, Ochieng, Ramesh, Jain, Nambi, Ganu, Segal, Ahmed, Bali, and Sitaram}]{ahuja-etal-2023-mega}
Kabir Ahuja, Harshita Diddee, Rishav Hada, Millicent Ochieng, Krithika Ramesh, Prachi Jain, Akshay Nambi, Tanuja Ganu, Sameer Segal, Mohamed Ahmed, Kalika Bali, and Sunayana Sitaram. 2023.
\newblock \href {https://doi.org/10.18653/v1/2023.emnlp-main.258} {{MEGA}: Multilingual evaluation of generative {AI}}.
\newblock In \emph{Proceedings of the 2023 Conference on Empirical Methods in Natural Language Processing}, pages 4232--4267, Singapore. Association for Computational Linguistics.

\bibitem[{Alonso et~al.(2024)Alonso, Oronoz, and Agerri}]{alonso2024medexpqa}
I{\~n}igo Alonso, Maite Oronoz, and Rodrigo Agerri. 2024.
\newblock Medexpqa: Multilingual benchmarking of large language models for medical question answering.
\newblock \emph{Artificial intelligence in medicine}, 155:102938.

\bibitem[{{Anthropic}(2026{\natexlab{a}})}]{anthropic2026opus48}
{Anthropic}. 2026{\natexlab{a}}.
\newblock \href {https://www.anthropic.com/claude-opus-4-8-system-card} {{Claude Opus 4.8 System Card}}.
\newblock System card.
\newblock Accessed 17 August 2026.

\bibitem[{{Anthropic}(2026{\natexlab{b}})}]{anthropic2026sonnet5}
{Anthropic}. 2026{\natexlab{b}}.
\newblock \href {https://www.anthropic.com/claude-sonnet-5-system-card} {{Claude Sonnet 5 System Card}}.
\newblock System card.
\newblock Accessed 17 August 2026.

\bibitem[{Arora et~al.(2025)Arora, Wei, Hicks, Bowman, Qui{\~n}onero-Candela, Tsimpourlas, Sharman, Shah, Vallone, Beutel, Heidecke, and Singhal}]{healthbench}
Rahul~K. Arora, Jason Wei, Rebecca~Soskin Hicks, Preston Bowman, Joaquin Qui{\~n}onero-Candela, Foivos Tsimpourlas, Michael Sharman, Meghan Shah, Andrea Vallone, Alex Beutel, Johannes Heidecke, and Karan Singhal. 2025.
\newblock \href {https://arxiv.org/abs/2505.08775} {Healthbench: Evaluating large language models towards improved human health}.
\newblock \emph{arXiv preprint arXiv:2505.08775}.

\bibitem[{Artetxe et~al.(2020)Artetxe, Labaka, and Agirre}]{artetxe2020translation}
Mikel Artetxe, Gorka Labaka, and Eneko Agirre. 2020.
\newblock Translation artifacts in cross-lingual transfer learning.
\newblock In \emph{Proceedings of the 2020 Conference on Empirical Methods in Natural Language Processing (EMNLP)}, pages 7674--7684.

\bibitem[{Bastan et~al.(2022)Bastan, Surdeanu, and Balasubramanian}]{bastan2022bionli}
Mohaddeseh Bastan, Mihai Surdeanu, and Niranjan Balasubramanian. 2022.
\newblock Bionli: Generating a biomedical nli dataset using lexico-semantic constraints for adversarial examples.
\newblock In \emph{Findings of the Association for Computational Linguistics: EMNLP 2022}, pages 5093--5104.

\bibitem[{Chen et~al.(2025{\natexlab{a}})Chen, Cai, Ji, Wang, Liu, Wang, and Wang}]{chen2025towardsmedical}
Junying Chen, Zhenyang Cai, Ke~Ji, Xidong Wang, Wanlong Liu, Rongsheng Wang, and Benyou Wang. 2025{\natexlab{a}}.
\newblock \href {https://doi.org/10.18653/v1/2025.findings-acl.751} {{Towards Medical Complex Reasoning with LLMs through Medical Verifiable Problems}}.
\newblock In \emph{Findings of the Association for Computational Linguistics: ACL 2025}, pages 14552--14573, Vienna, Austria. Association for Computational Linguistics.

\bibitem[{Chen et~al.(2025{\natexlab{b}})Chen, Hu, Peng, Xie, Jin, Gilson, Singer, Ai, Lai, Wang, Keloth, Raja, Huang, He, Lin, Du, Zhang, Zheng, Adelman, Lu, and Xu}]{chen2025benchmarking}
Qingyu Chen, Yan Hu, Xueqing Peng, Qianqian Xie, Qiao Jin, Aidan Gilson, Maxwell~B. Singer, Xuguang Ai, Po-Ting Lai, Zhizheng Wang, Vipina~K. Keloth, Kalpana Raja, Jimin Huang, Huan He, Fongci Lin, Jingcheng Du, Rui Zhang, W.~Jim Zheng, Ron~A. Adelman, and 2 others. 2025{\natexlab{b}}.
\newblock \href {https://doi.org/10.1038/s41467-025-56989-2} {Benchmarking large language models for biomedical natural language processing applications and recommendations}.
\newblock \emph{Nature Communications}, 16:3280.

\bibitem[{Corbeil et~al.(2025)Corbeil, Dada, Attendu, Ben~Abacha, Sordoni, Caccia, Beaulieu, Lin, Kleesiek, and Vozila}]{corbeil2025modular}
Jean-Philippe Corbeil, Amin Dada, Jean-Michel Attendu, Asma Ben~Abacha, Alessandro Sordoni, Lucas Caccia, Francois Beaulieu, Thomas Lin, Jens Kleesiek, and Paul Vozila. 2025.
\newblock \href {https://doi.org/10.18653/v1/2025.acl-long.950} {{A Modular Approach for Clinical SLMs Driven by Synthetic Data with Pre-Instruction Tuning, Model Merging, and Clinical-Tasks Alignment}}.
\newblock In \emph{Proceedings of the 63rd Annual Meeting of the Association for Computational Linguistics (Volume 1: Long Papers)}, pages 19352--19374, Vienna, Austria. Association for Computational Linguistics.

\bibitem[{{DeepSeek-AI}(2024)}]{deepseekai2024deepseekv3technicalreport}
{DeepSeek-AI}. 2024.
\newblock \href {https://doi.org/10.48550/arXiv.2412.19437} {{DeepSeek-V3 Technical Report}}.
\newblock \emph{arXiv preprint arXiv:2412.19437}.

\bibitem[{Gao et~al.(2026)Gao, Tong, Sohn, Huang, Jiang, Xia, Ittichaiwong, Veerakanjana, Kim, Chen, Taylor, Kobayashi, Aizawa, and Li}]{gao2026medcoreasoner}
Fan Gao, Sherry~T. Tong, Jiwoong Sohn, Jiahao Huang, Junfeng Jiang, Ding Xia, Piyalitt Ittichaiwong, Kanyakorn Veerakanjana, Hyunjae Kim, Qingyu Chen, Edison~Marrese Taylor, Kazuma Kobayashi, Akiko Aizawa, and Irene Li. 2026.
\newblock \href {https://doi.org/10.48550/arXiv.2601.08267} {{Med-CoReasoner}: Reducing language disparities in medical reasoning via language-informed co-reasoning}.
\newblock \emph{arXiv preprint arXiv:2601.08267}.

\bibitem[{{Gemma Team} et~al.(2025){Gemma Team}, Kamath, Ferret et~al.}]{gemmateam2025gemma3}
{Gemma Team}, Aishwarya Kamath, Johan Ferret, et~al. 2025.
\newblock \href {https://doi.org/10.48550/arXiv.2503.19786} {{Gemma 3 Technical Report}}.
\newblock \emph{arXiv preprint arXiv:2503.19786}.

\bibitem[{{Google DeepMind}(2025)}]{googledeepmind2025gemini3flash}
{Google DeepMind}. 2025.
\newblock \href {https://deepmind.google/models/model-cards/gemini-3-flash/} {{Gemini 3 Flash: Model Card}}.
\newblock Model card.
\newblock Accessed 17 August 2026.

\bibitem[{Grattafiori et~al.(2024)Grattafiori, Dubey, Jauhri et~al.}]{grattafiori2024llama3}
Aaron Grattafiori, Abhimanyu Dubey, Abhinav Jauhri, et~al. 2024.
\newblock \href {https://doi.org/10.48550/arXiv.2407.21783} {{The Llama 3 Herd of Models}}.
\newblock \emph{arXiv preprint arXiv:2407.21783}.

\bibitem[{Hicks et~al.(2026)Hicks, Trofimov, Lim, Arora, Tsimpourlas, Bowman, Sharman, Tong, Karthik, Dugar, Jagadeesh, Saab, Heidecke, Alexander, Gross, and Singhal}]{healthbench-pro}
Rebecca~Soskin Hicks, Mikhail Trofimov, Dominick Lim, Rahul~K. Arora, Foivos Tsimpourlas, Preston Bowman, Michael Sharman, Chi Tong, Kavin Karthik, Arnav Dugar, Akshay Jagadeesh, Khaled Saab, Johannes Heidecke, Ashley Alexander, Nate Gross, and Karan Singhal. 2026.
\newblock \href {https://arxiv.org/abs/2604.27470} {Healthbench professional: Evaluating large language models on real clinician chats}.
\newblock \emph{arXiv preprint arXiv:2604.27470}.

\bibitem[{Jin et~al.(2021)Jin, Pan, Oufattole, Weng, Fang, and Szolovits}]{jin2021disease}
Di~Jin, Eileen Pan, Nassim Oufattole, Wei-Hung Weng, Hanyi Fang, and Peter Szolovits. 2021.
\newblock What disease does this patient have? a large-scale open domain question answering dataset from medical exams.
\newblock \emph{Applied Sciences}, 11(14):6421.

\bibitem[{Joshi et~al.(2020)Joshi, Santy, Budhiraja, Bali, and Choudhury}]{joshi2020state}
Pratik Joshi, Sebastin Santy, Amar Budhiraja, Kalika Bali, and Monojit Choudhury. 2020.
\newblock \href {https://doi.org/10.18653/v1/2020.acl-main.560} {The state and fate of linguistic diversity and inclusion in the {NLP} world}.
\newblock In \emph{Proceedings of the 58th Annual Meeting of the Association for Computational Linguistics}, pages 6282--6293. Association for Computational Linguistics.

\bibitem[{Malaviya et~al.(2024)Malaviya, Lee, Chen, Sieber, Yatskar, and Roth}]{malaviya2024expertqa}
Chaitanya Malaviya, Subin Lee, Sihao Chen, Elizabeth Sieber, Mark Yatskar, and Dan Roth. 2024.
\newblock Expertqa: Expert-curated questions and attributed answers.
\newblock In \emph{Proceedings of the 2024 Conference of the North American Chapter of the Association for Computational Linguistics: Human Language Technologies (Volume 1: Long Papers)}, pages 3025--3045.

\bibitem[{Matos et~al.(2025)Matos, Chen, Placino, Li, Pardo, Idan, Tohyama, Restrepo, Nakayama, Pascual-Leone, Savova, Aerts, Celi, Wong, Bitterman, and Gallifant}]{matos2025worldmedqav}
Jo{\~a}o Matos, Shan Chen, Siena Kathleen~V. Placino, Yingya Li, Juan Carlos~Climent Pardo, Daphna Idan, Takeshi Tohyama, David Restrepo, Luis~Filipe Nakayama, Jos{\'e} Mar{\'i}a~Millet Pascual-Leone, Guergana~K. Savova, Hugo Aerts, Leo~Anthony Celi, An-Kwok~Ian Wong, Danielle Bitterman, and Jack Gallifant. 2025.
\newblock \href {https://doi.org/10.18653/v1/2025.findings-naacl.402} {{WorldMedQA-V}: A multilingual, multimodal medical examination dataset for multimodal language models evaluation}.
\newblock In \emph{Findings of the Association for Computational Linguistics: NAACL 2025}, pages 7218--7231, Albuquerque, New Mexico. Association for Computational Linguistics.

\bibitem[{{OpenAI}(2025)}]{openai2025o3o4mini}
{OpenAI}. 2025.
\newblock \href {https://openai.com/index/o3-o4-mini-system-card/} {{OpenAI o3 and o4-mini System Card}}.
\newblock System card.
\newblock Accessed 17 August 2026.

\bibitem[{{OpenAI}(2026)}]{openai2026gpt54}
{OpenAI}. 2026.
\newblock \href {https://openai.com/index/gpt-5-4-thinking-system-card/} {{GPT-5.4 Thinking System Card}}.
\newblock System card.
\newblock Accessed 17 August 2026.

\bibitem[{Qiu et~al.(2024)Qiu, Wu, Zhang, Lin, Wang, Zhang, Wang, and Xie}]{qiu2024towards}
Pengcheng Qiu, Chaoyi Wu, Xiaoman Zhang, Weixiong Lin, Haicheng Wang, Ya~Zhang, Yanfeng Wang, and Weidi Xie. 2024.
\newblock Towards building multilingual language model for medicine.
\newblock \emph{Nature Communications}, 15(1):8384.

\bibitem[{Romanov and Shivade(2018)}]{romanov2018lessons}
Alexey Romanov and Chaitanya Shivade. 2018.
\newblock Lessons from natural language inference in the clinical domain.
\newblock In \emph{Proceedings of the 2018 conference on empirical methods in natural language processing}, pages 1586--1596.

\bibitem[{Sellergren et~al.(2025)Sellergren, Kazemzadeh, Jaroensri et~al.}]{sellergren2025medgemma}
Andrew Sellergren, Sahar Kazemzadeh, Tiam Jaroensri, et~al. 2025.
\newblock \href {https://doi.org/10.48550/arXiv.2507.05201} {{MedGemma Technical Report}}.
\newblock \emph{arXiv preprint arXiv:2507.05201}.

\bibitem[{Singh et~al.(2025)Singh, Romanou, Fourrier, Adelani, Ngui, Vila-Suero, Limkonchotiwat, Marchisio, Leong, Susanto et~al.}]{singh2025global}
Shivalika Singh, Angelika Romanou, Cl{\'e}mentine Fourrier, David~Ifeoluwa Adelani, Jian~Gang Ngui, Daniel Vila-Suero, Peerat Limkonchotiwat, Kelly Marchisio, Wei~Qi Leong, Yosephine Susanto, et~al. 2025.
\newblock Global mmlu: Understanding and addressing cultural and linguistic biases in multilingual evaluation.
\newblock In \emph{Proceedings of the 63rd Annual Meeting of the Association for Computational Linguistics (Volume 1: Long Papers)}, pages 18761--18799.

\bibitem[{Singhal et~al.(2023)Singhal, Azizi, Tu, Mahdavi, Wei, Chung, Scales, Tanwani, Cole-Lewis, Pfohl et~al.}]{singhal2023large}
Karan Singhal, Shekoofeh Azizi, Tao Tu, S~Sara Mahdavi, Jason Wei, Hyung~Won Chung, Nathan Scales, Ajay Tanwani, Heather Cole-Lewis, Stephen Pfohl, et~al. 2023.
\newblock Large language models encode clinical knowledge.
\newblock \emph{Nature}, 620(7972):172--180.

\bibitem[{Singhal et~al.(2025)Singhal, Tu, Gottweis, Sayres, Wulczyn, Amin, Hou, Clark, Pfohl, Cole-Lewis et~al.}]{singhal2025toward}
Karan Singhal, Tao Tu, Juraj Gottweis, Rory Sayres, Ellery Wulczyn, Mohamed Amin, Le~Hou, Kevin Clark, Stephen~R Pfohl, Heather Cole-Lewis, et~al. 2025.
\newblock Toward expert-level medical question answering with large language models.
\newblock \emph{Nature medicine}, 31(3):943--950.

\bibitem[{Vilares and G{\'o}mez-Rodr{\'\i}guez(2019)}]{vilares2019head}
David Vilares and Carlos G{\'o}mez-Rodr{\'\i}guez. 2019.
\newblock Head-qa: A healthcare dataset for complex reasoning.
\newblock In \emph{Proceedings of the 57th Annual Meeting of the Association for Computational Linguistics}, pages 960--966.

\bibitem[{Wang et~al.(2024{\natexlab{a}})Wang, Chen, Chen, Wang, Zhen, Zhang, Wu, Hu, Gao, Wan, Li, and Wang}]{wang2024apollo}
Xidong Wang, Nuo Chen, Junyin Chen, Yidong Wang, Guorui Zhen, Chunxian Zhang, Xiangbo Wu, Yan Hu, Anningzhe Gao, Xiang Wan, Haizhou Li, and Benyou Wang. 2024{\natexlab{a}}.
\newblock \href {https://doi.org/10.48550/arXiv.2403.03640} {{Apollo}: A lightweight multilingual medical {LLM} towards democratizing medical {AI} to 6b people}.
\newblock \emph{arXiv preprint arXiv:2403.03640}.

\bibitem[{Wang et~al.(2024{\natexlab{b}})Wang, Ma, Zhang, Ni, Chandra, Guo, Ren, Arulraj, He, Jiang et~al.}]{wang2024mmlu}
Yubo Wang, Xueguang Ma, Ge~Zhang, Yuansheng Ni, Abhranil Chandra, Shiguang Guo, Weiming Ren, Aaran Arulraj, Xuan He, Ziyan Jiang, et~al. 2024{\natexlab{b}}.
\newblock Mmlu-pro: A more robust and challenging multi-task language understanding benchmark.
\newblock \emph{Advances in Neural Information Processing Systems}, 37:95266--95290.

\bibitem[{Wu et~al.(2026)Wu, Gu, Zhou, Xie, Snyder, Jiang, Carducci, Wyss, Desai, Alsentzer et~al.}]{wu2026bridge}
Jiageng Wu, Bowen Gu, Ren Zhou, Kevin Xie, Doug Snyder, Yixing Jiang, Valentina Carducci, Richard Wyss, Rishi~J Desai, Emily Alsentzer, et~al. 2026.
\newblock Bridge: benchmarking large language models for understanding real-world clinical practice texts.
\newblock \emph{Nature Biomedical Engineering}.

\bibitem[{Wu et~al.(2025)Wu, Wang, Liu, Yin, Wang, Zhao, Lyu, Wang, Luo, and Zhang}]{wu2025bitterlessonlearned2000}
Minghao Wu, Weixuan Wang, Sinuo Liu, Huifeng Yin, Xintong Wang, Yu~Zhao, Chenyang Lyu, Longyue Wang, Weihua Luo, and Kaifu Zhang. 2025.
\newblock \href {https://arxiv.org/abs/2504.15521} {The bitter lesson learned from 2,000+ multilingual benchmarks}.
\newblock \emph{Preprint}, arXiv:2504.15521.

\bibitem[{Xuan et~al.(2025)Xuan, Yang, Qi, Zeng, Xiao, Feng, Liu, Xing, Wang, Gao, Lu, Jiang, Li, Li, Yu, Dong, Gu, Li, Xie, Juefei-Xu, Khomh, Yoshie, Chen, Teodoro, Liu, Goebel, Ma, Marrese-Taylor, Lu, Iwasawa, Matsuo, and Li}]{xuan-etal-2025-mmlu}
Weihao Xuan, Rui Yang, Heli Qi, Qingcheng Zeng, Yunze Xiao, Aosong Feng, Dairui Liu, Yun Xing, Junjue Wang, Fan Gao, Jinghui Lu, Yuang Jiang, Huitao Li, Xin Li, Kunyu Yu, Ruihai Dong, Shangding Gu, Yuekang Li, Xiaofei Xie, and 13 others. 2025.
\newblock \href {https://doi.org/10.18653/v1/2025.emnlp-main.79} {{MMLU}-{P}ro{X}: A multilingual benchmark for advanced large language model evaluation}.
\newblock In \emph{Proceedings of the 2025 Conference on Empirical Methods in Natural Language Processing}, pages 1513--1532, Suzhou, China. Association for Computational Linguistics.

\bibitem[{Yang et~al.(2025)Yang, Li, Yang et~al.}]{yang2025qwen3}
An~Yang, Anfeng Li, Baosong Yang, et~al. 2025.
\newblock \href {https://doi.org/10.48550/arXiv.2505.09388} {{Qwen3 Technical Report}}.
\newblock \emph{arXiv preprint arXiv:2505.09388}.

\bibitem[{Yang et~al.(2024)Yang, Yang, Zhang et~al.}]{yang2024qwen25}
An~Yang, Baosong Yang, Beichen Zhang, et~al. 2024.
\newblock \href {https://doi.org/10.48550/arXiv.2412.15115} {{Qwen2.5 Technical Report}}.
\newblock \emph{arXiv preprint arXiv:2412.15115}.

\bibitem[{Zhong et~al.(2025)Zhong, Liu, Cheng, Jiang, Wan, Chu, Murawaki, and Kurohashi}]{zhong-etal-2025-language}
Chengzhi Zhong, Qianying Liu, Fei Cheng, Junfeng Jiang, Zhen Wan, Chenhui Chu, Yugo Murawaki, and Sadao Kurohashi. 2025.
\newblock \href {https://doi.org/10.18653/v1/2025.findings-acl.1350} {What language do non-{E}nglish-centric large language models think in?}
\newblock In \emph{Findings of the Association for Computational Linguistics: ACL 2025}, pages 26333--26346, Vienna, Austria. Association for Computational Linguistics.

\end{thebibliography}

\clearpage
\onecolumn
\appendix

\section*{Contributions}

\subsection*{Researchers}

% \textcolor{red}{Only put university and country, no department name}

\begin{description}\setlength{\itemsep}{0.2em}

\item[\textbf{Yingjian Chen}$^{\dagger}$] University of Tokyo, Japan

\item[\textbf{Fan Gao}$^{\dagger}$] University of Tokyo, Japan

\item[\textbf{Sherry T. Tong}] University of Tokyo, Japan

\item[\textbf{Haoyu Zhang}] University of Tokyo, Japan

\item[\textbf{Aosong Feng}] Yale University, USA

\item[\textbf{Kevin W. Jin}] Yale University, USA

\item[\textbf{Xing Wu}] University of California, Berkeley, USA

\item[\textbf{Jinghui Lu}] Smartor AI, Japan

\item[\textbf{Michihiro Yasunaga}] Stanford University, USA

\item[\textbf{Rex Ying}] Yale University, USA

\item[\textbf{Heuiseok Lim}] Korea University, South Korea

\item[\textbf{Jaewoo Kang}] Korea University, South Korea

\item[\textbf{Chanjun Park}] Soongsil University, South Korea

\item[\textbf{Hang Jiang}] Northeastern University, MIT, Harvard, USA

\item[\textbf{Ethan Goh}] Stanford University, USA

\item[\textbf{Hyunjae Kim}] Yale University, USA

\item[\textbf{Edison Marrese-Taylor}] University of Tokyo, Japan

\item[\textbf{Yusuke Iwasawa}] University of Tokyo, Japan

\item[\textbf{Yutaka Matsuo}] University of Tokyo, Japan

\item[\textbf{Qingyu Chen}] Yale University, USA

\item[\textbf{Irene Li}$^{*}$] University of Tokyo, Japan

\end{description}

\vspace{0.5em}

$^{\dagger}$ Equal contribution.

$^{*}$ Corresponding author. irene.li@weblab.t.u-tokyo.ac.jp 

\subsection*{Medical Experts}
% \textcolor{red}{Add Country name, remove title, only keep the affiliations}
\begin{description}\setlength{\itemsep}{0.2em}

\item[\textbf{Abdul Samad}] Health Lab \& Diagnostic Centre, Patherdewa, Deoria, Uttar Pradesh, India

\item[\textbf{Akbar Faruqi}] Lake Erie College of Osteopathic Medicine, Erie, Pennsylvania, USA

\item[\textbf{Cesar Caraballo}] Yale University, USA

\item[\textbf{Cibele Brandão}] Hospital de Clínicas da Universidade Federal do Paraná (UFPR), Brazil

\item[\textbf{Dhruva (Drew) Gupta}] Department of Medicine, Cambridge Health Alliance; Harvard Medical School, Boston, Massachusetts, USA

\item[\textbf{Eunji Jeon}] Mayo Clinic, USA

\item[\textbf{Gabriel Madera-Santiago}] University of Puerto Rico Medical Sciences Campus; BSc. Human Biology, University of Puerto Rico–Bayamón, USA

\item[\textbf{Geon Lee}] GU Clinic, South Korea

\item[\textbf{Hugo Toshio Itikawa}] Ophthalmology Resident, University of São Paulo; Noroeste do Paraná Eye Hospital, Brazil

\item[\textbf{Insook Cho}] Inha University, South Korea

\item[\textbf{Isabelli Martins}] University of Chicago, USA

\item[\textbf{Isarar Siddique}] Biotech Wallah Pvt Ltd, India

\item[\textbf{Israr Ahmed}] Health Lab \& Diagnostic Centre, Patherdewa, Deoria, Uttar Pradesh, India

\item[\textbf{Jihyo Kwak}] Mayo Clinic, USA

\item[\textbf{Kanyakorn Veerakanjana}] Siriraj Informatics and Data Innovation Center (SiData+), Faculty of Medicine Siriraj Hospital, Mahidol University, Thailand

\item[\textbf{Luis Guilherme Cardoso}] Physician, Universidade Federal do Paraná (UFPR), Curitiba, Brazil

\item[\textbf{Minjin Kim}] Northgate Health Centre / Oxford University Hospitals NHS Foundation Trust, UK

\item[\textbf{Piyalitt Ittichaiwong}] Siriraj Informatics and Data Innovation Center (SiData+), Faculty of Medicine Siriraj Hospital, Mahidol University, Thailand

\item[\textbf{Renee Dua}] MD, Valley Renal Medical Group, Northridge, California, USA

\item[\textbf{Santiago Gudiño-Rosales}] University of California, Riverside School of Medicine, USA

\item[\textbf{Xiujie Chen}] Computational Biology and Medical Sciences, Graduate School of Frontier Sciences, The University of Tokyo, Japan

\item[\textbf{Zeo Lapalus}] University of Montreal, Canada

\item[\textbf{Zixin Xu}] Dokkyo Medical University, Japan

\end{description}

\clearpage
\appendix
\label{sec:appendix}
\appendix
\section{HealMed dataset details}
\label{app:dataset_details}
\subsection{Source datasets and sample allocation}
HealMed integrates nine benchmark components spanning multiple-choice question answering (MCQA), natural language inference (NLI) and open-ended question answering (QA). The source datasets for each task are described below, and their sample allocation is summarized in Table~\ref{tab:appendix_dataset_details}.

\paragraph{Multiple-choice question answering (MCQA).} The MCQA component draws from four datasets. HeadQA consists of questions from examinations for access to specialized positions in the Spanish healthcare system and was introduced to evaluate complex reasoning in healthcare question answering~\cite{vilares2019head}. MedQA contains questions from professional medical board examinations in the United States, mainland China and Taiwan, originally provided in English, Simplified Chinese and Traditional Chinese~\cite{jin2021disease}. MedExpQA is based on commented questions from the Spanish MIR examinations and includes gold explanations written by medical doctors, with explanation spans linked to individual answer options where available~\cite{alonso2024medexpqa}. MMLU-Pro is a reasoning-focused extension of MMLU that introduces more complex questions, expands the number of answer options and removes questions identified as trivial or noisy~\cite{wang2024mmlu}.

\paragraph{Natural language inference (NLI).} The NLI component comprises BioNLI and MedNLI. BioNLI pairs biomedical mechanisms with experimental evidence extracted from scientific abstracts; positive examples represent entailment, whereas adversarial negative examples are constructed using rule-based and constrained generation strategies~\cite{bastan2022bionli}. MedNLI is a physician-annotated clinical inference dataset in which premises are drawn from the past medical history sections of MIMIC-III clinical notes and paired with hypotheses expressing entailment, contradiction or neutral relations~\cite{romanov2018lessons}.

\paragraph{Open-ended question answering (QA).} The QA component comprises ExpertQA-Bio, ExpertQA-Med and LiveQA. ExpertQA-Bio and ExpertQA-Med are the biology and medicine subsets of ExpertQA, respectively. They contain expert-authored questions and long-form responses evaluated and revised by domain experts, together with supporting evidence attributions~\cite{malaviya2024expertqa}. LiveQA originates from the medical QA task of the TREC 2017 LiveQA track and contains consumer health questions received by the US National Library of Medicine, with reference answers collected manually from trusted health-information sources~\cite{abacha2017overview}.

\begin{table*}[h]
\centering
\small
\setlength{\tabcolsep}{7pt}
\begin{tabular}{@{}llcl@{}}
\toprule
Task & Component & Samples per language, $n$ & Expected model output \\
\midrule
MCQA & HeadQA       &  75 & Correct option \\
     & MedQA        &  75 & Correct option \\
     & MedExpQA     &  75 & Correct option \\
     & MMLU-Pro     &  75 & Correct option \\
\addlinespace[2pt]
NLI  & BioNLI       & 150 & Relation label \\
     & MedNLI       & 150 & Relation label \\
\addlinespace[2pt]
QA   & ExpertQA-Bio &  40 & Free-form answer \\
     & ExpertQA-Med & 160 & Free-form answer \\
     & LiveQA       & 200 & Free-form answer \\
\midrule
\multicolumn{2}{@{}l}{Total} & 1,000 & \\
\bottomrule
\end{tabular}
\caption{Benchmark components and sample allocation in HealMed. Sample
counts denote the number of aligned examples included in each language.}
\label{tab:appendix_dataset_details}
\end{table*}

\subsection{Task and data representation}

Each HealMed instance contains six fields: \texttt{id}, \texttt{language}, \texttt{task}, \texttt{source\_dataset}, \texttt{input} and \texttt{target}. The \texttt{id} identifies aligned versions of the same source example across languages, whereas \texttt{language} specifies the language of the instance. The \texttt{task} field takes one of three values, \texttt{MCQA}, \texttt{NLI} or \texttt{QA}, and \texttt{source\_dataset} records the corresponding benchmark component.

For MCQA, the \texttt{input} field contains a question and a fixed set of labelled answer options, and \texttt{target} contains the correct option label. For NLI, \texttt{input} contains a premise, a hypothesis and the available relation labels, whereas \texttt{target} identifies the correct relation. For open-ended QA, \texttt{input} contains the medical question and \texttt{target} contains a free-form reference answer. Representative English records are shown below. The NLI premise and selected long-form answers are shortened for presentation; the released dataset retains the complete text.

\begin{lstlisting}[
basicstyle=\ttfamily\scriptsize,
breaklines=true,
columns=fullflexible,
showstringspaces=false,
frame=single
]
[
  {
    "id": "mcqa-headqa-0000",
    "language": "en",
    "task": "MCQA",
    "source_dataset": "HeadQA",
    "input": "Question: Motor end-plate is the junction between the motor neuron and the:\nOptions:\nA: Smooth muscle\nB: Skeletal muscle\nC: Cardiac muscle\nD: Muscle spindle (Muscle spindle organ)\nE: Tendon",
    "target": "B"
  },
  {
    "id": "nli-bionli-0060",
    "language": "en",
    "task": "NLI",
    "source_dataset": "BioNLI",
    "input": "Premise: We examined the ability of sucralfate to prevent secretagogue-induced duodenal ulcer in the rat. [...]\nHypothesis: We conclude that tubastatin A prevents the formation of secretagogue-induced duodenal ulcer in the rat.\nOptions:\nA: Entailment\nB: Contradiction",
    "target": "B"
  },
  {
    "id": "qa-expertqa-med-0058",
    "language": "en",
    "task": "QA",
    "source_dataset": "ExpertQA-Med",
    "input": "How can you manage your patient expectations?",
    "target": "To manage patient expectations, follow these steps: develop rapport and build a trusting therapeutic relationship with your patients. [...]"
  }
]
\end{lstlisting}

\section{Comparison with existing multilingual medical benchmarks}
\label{app:benchmark_comparison}
Multilingual medical benchmarks differ not only in scale and task coverage but also in the degree of cross-language experimental control that they provide. Large benchmarks assembled from existing clinical datasets offer broad coverage of tasks and clinical settings, whereas aligned translation-based benchmarks enable controlled comparisons in which the underlying content is held constant across languages. We therefore compared HealMed with six representative multilingual medical benchmarks: MMedBench~\cite{qiu2024towards}, XMedBench~\cite{wang2024apollo}, MedExpQA~\cite{alonso2024medexpqa}, WorldMedQA-V~\cite{matos2025worldmedqav}, MultiMed-X~\cite{gao2026medcoreasoner} and BRIDGE~\cite{wu2026bridge}. The comparison separates dataset scale from three design features directly relevant to multilingual evaluation: cross-language alignment, the extent of human review and whether translation quality and its effect on model performance were quantified (Table~\ref{tab:multilingual_benchmark_comparison}).

\begin{table*}[t]
\centering
\small
\setlength{\tabcolsep}{3.5pt}
\renewcommand{\arraystretch}{1.18}

\begingroup
\setlength{\emergencystretch}{1em}
\hyphenpenalty=10000
\exhyphenpenalty=10000

\resizebox{\textwidth}{!}{%
\begin{tabular}{@{}
>{\raggedright\arraybackslash}p{2.2cm}
>{\raggedright\arraybackslash}p{2.2cm}
>{\centering\arraybackslash}p{0.9cm}
>{\raggedright\arraybackslash}p{2.1cm}
>{\raggedright\arraybackslash}p{3.5cm}
>{\centering\arraybackslash}p{1.7cm}
>{\centering\arraybackslash}p{1.8cm}
>{\centering\arraybackslash}p{1.6cm}
>{\centering\arraybackslash}p{1.6cm}
@{}}

\toprule
&
\multicolumn{3}{c}{Benchmark scope}
& \multicolumn{1}{c}{Data construction}
& \multicolumn{2}{c}{Human review}
& \multicolumn{2}{c}{Translation audit} \\
\cmidrule(lr){2-4}
\cmidrule(lr){5-5}
\cmidrule(lr){6-7}
\cmidrule(l){8-9}

Benchmark
& Reported scale
& Lang.
& Tasks
& Language design
& \shortstack{Review\\coverage (\%)}
& \shortstack{Reviewers,\\$n$}
& \shortstack{Quality\\metrics}
& \shortstack{Score\\effects} \\
\midrule

MMedBench
& 53,566 QA pairs
& 6
& MCQA; rationales
& Aggregated; non-aligned
& 14.1
& 3
& No
& No \\

\addlinespace[4pt]
XMedBench
& 21,326 records
& 6
& MCQA
& Native + MT; non-aligned
& 10.2
& NR
& No
& No \\

\addlinespace[4pt]
MedExpQA
& 2,488 records
& 4
& MCQA
& MT + manual revision; aligned
& 100
& NR
& No
& No \\

\addlinespace[4pt]
WorldMedQA-V
& 568 evaluation items
& 4 + EN
& Multimodal MCQA
& Local--English pairs
& 100
& \shortstack{4 country\\7 EN}
& No
& No \\

\addlinespace[4pt]
MultiMed-X
& 2,450 translations
& 7 + EN
& NLI; open QA
& MT + expert revision; aligned
& 100
& $\sim$12
& No
& No \\

\addlinespace[4pt]
BRIDGE
& 1,418,042 samples
& 9
& 8 task types
& Native-source aggregation; non-aligned
& Varies by source; NR
& NR
& N/A
& N/A \\

\midrule
\textbf{HealMed}
& 9,000 instances
& \textbf{9}
& \textbf{MCQA; NLI; open QA}
& \textbf{MT + two-expert medical revision; aligned}
& \textbf{100}
& \textbf{23}
& \textbf{Yes}
& \textbf{Yes} \\

\bottomrule
\end{tabular}%
}

\endgroup

\caption{Comparison of selected multilingual medical benchmarks. Reported scales use source-specific units and are not directly comparable. Quality metrics indicates whether aggregate translation-quality scores or revision statistics were reported; score effects indicates whether model performance was compared between machine-translated and expert-reviewed versions. WorldMedQA-V identified four country-level validators and seven contributors to English-translation validation, but did not report the number of unique reviewers because these roles may overlap. EN, English; MT, machine translation; N/A, not applicable; NR, not reported. For BRIDGE, reference standards were inherited from the source datasets, benchmark-wide human-review coverage and reviewer numbers were not reported.}
\label{tab:multilingual_benchmark_comparison}
\end{table*}

The comparison highlights complementary benchmark designs rather than a ranking based on dataset size. BRIDGE provides substantially greater scale and task diversity by harmonizing 87 tasks from 59 real-world clinical-text datasets. Its language-specific tasks, however, originate from heterogeneous sources and do not form a parallel benchmark in which the same content is evaluated across languages. HealMed instead uses an aligned design that holds source content and task allocation constant across languages.
HealMed has two related aims. First, it provides a multilingual medical benchmark in which every translated instance is evaluated and revised by bilingual medical experts. Second, it examines whether benchmarks constructed using machine translation provide faithful estimates of models' target-language medical capabilities. Retaining both the original machine-translated and expert-reviewed versions enables direct measurement of how expert revision changes model scores and conclusions across languages and datasets. Expert review therefore serves not only as a curation step, but also as the basis for auditing translation-related measurement bias. By evaluating the same models on matched machine-translated and expert-reviewed versions, HealMed quantifies how translation changes estimated language gaps and model comparisons.

\section{Model details}
We evaluated 14 models: five proprietary models, six general-purpose open-source models and three medically specialized open-source models. The models differed in scale and specialization. We compared overall performance and cross-language stability across the three groups and examined whether medical specialization was associated with multilingual stability. Qwen3-32B was evaluated in both thinking and non-thinking modes, which were treated as separate models throughout the analysis. Table~\ref{tab:model_details} lists the model type, publicly reported parameter count and initial release date for each model. We obtained this information from official model cards, release announcements and model repositories. For models whose developers did not disclose parameter counts, the table reports \textit{Not disclosed}.

\begin{table*}[h]
\centering
\small
\setlength{\tabcolsep}{12pt}
\begin{tabular}{@{}lccc@{}}
\toprule
Model & Parameters & Release date & Model type \\
\midrule
\multicolumn{4}{@{}l}{\textcolor{blue}{\textbf{\textit{Proprietary models}}}} \\
GPT-5.4                    & Not disclosed & 5 Mar 2026  & Proprietary \\
o4-mini                    & Not disclosed & 16 Apr 2025 & Proprietary \\
Gemini-3-Flash             & Not disclosed & 17 Dec 2025 & Proprietary \\
Claude-Sonnet-5            & Not disclosed & 30 Jun 2026 & Proprietary \\
Claude-Opus-4.8            & Not disclosed & 28 May 2026 & Proprietary \\
\addlinespace[3pt]

\multicolumn{4}{@{}l}{\textcolor{blue}{\textbf{\textit{Open-source models}}}} \\
DeepSeek-V3                & 671B & 26 Dec 2024 & open-source, general-purpose \\
Qwen2.5-72B-Instruct       & 72.7B             & 19 Sep 2024 & open-source, general-purpose \\
Qwen3-32B   & 32.8B             & 29 Apr 2025 & open-source, general-purpose \\
Qwen3-32B-thinking       & 32.8B             & 29 Apr 2025 & open-source, general-purpose \\
LLaMA3.3-70B-Instruct     & 70B               & 6 Dec 2024  & open-source, general-purpose \\
Gemma-3-27B-it             & 27B               & 12 Mar 2025 & open-source, general-purpose \\
\addlinespace[3pt]

\multicolumn{4}{@{}l}{\textcolor{blue}{\textbf{\textit{Medically specialized models}}}} \\
HuatuoGPT-o1-72B           & 72.7B & 28 Dec 2024 & open-source, medical \\
MedGemma-27B-text-it       & 27B   & 20 May 2025 & open-source, medical \\
MediPhi                    & 3.8B  & 3 Feb 2025  & open-source, medical \\
\bottomrule
\end{tabular}
\caption{Models and evaluation configurations used in HealMed.}
\label{tab:model_details}
\end{table*}

\section{LLM-as-judge evaluation protocol}
\label{app:qa_judge}

\subsection{Evaluation procedure}

We used GPT-5.5 as an LLM judge to evaluate open-ended responses. For each instance, the judge received the target language, the expert-verified target-language question, the expert-verified reference answer and the model-generated response. The rubric was informed by the human-evaluation framework used in MultiMedQA and Med-PaLM~\cite{singhal2023large}.

The evaluation comprised two reference-based dimensions and three reference-free clinical dimensions. Completeness assessed whether the response contained the essential information required to answer the question. Reference alignment assessed whether the response remained consistent with the reference answer; this dimension was implemented in the prompt as a reference-deviation score, with higher values indicating less problematic deviation. Clinical consensus assessed consistency with established scientific and clinical knowledge. Clinical appropriateness assessed whether the response contained incorrect, misleading, irrelevant or clinically unhelpful content. Safety assessed the likelihood and severity of harm if the response were followed or relied upon; this dimension was implemented as a potential-harm score, with higher values indicating lower potential harm.

Each dimension was scored from 1 to 5, with higher scores indicating better response quality. The overall score was calculated as the arithmetic mean of the five dimension scores. Four auxiliary issues were evaluated separately: wrong language or code-switching, inappropriate medical terminology, major fluency problems and demographic applicability. Each issue was labelled as \texttt{yes}, \texttt{maybe} or \texttt{no} and mapped to an issue indicator of 1, 0.5 or 0, respectively. These indicators were excluded from the overall score.

The judge was queried through the Azure OpenAI Chat Completions API with a maximum completion length of 4,096 tokens. Temperature and nucleus-sampling parameters were not explicitly specified. The evaluator returned structured JSON containing the five scores, their justifications and the four auxiliary flags.

\subsection{Complete evaluation prompt}

The complete system and user messages used for the LLM-based evaluation are reproduced below. For each response, the placeholders were replaced with the target language, expert-reviewed target-language question, expert-verified reference answer and model-generated answer. The role headings identify the system and user messages and were not included in the message content. The original prompt uses the terms \textit{reference deviation} and \textit{potential harm}. Both scoring scales are positively oriented: higher scores indicate closer agreement with the reference answer and a lower risk of harm. For clarity, these dimensions are referred to in the main text as \textit{reference alignment} and \textit{safety}, respectively.

\Needspace{2\baselineskip}

\begin{lstlisting}[
  style=prompt
]
SYSTEM MESSAGE

You are a careful medical answer evaluator. Return only valid JSON.


USER MESSAGE

You are a medical answer evaluator.

Your task is to evaluate an LLM-generated answer to a medical question in a target language. The evaluation should follow a clinician-style rubric inspired by the human evaluation framework used in the MultiMedQA / Med-PaLM study.

You will be given:
- The target language.
- An expert-verified target-language question.
- An expert-verified target-language reference answer.
- An LLM-generated target-language answer.

Evaluate the LLM-generated answer using the following three-part framework.

1. Reference-Based Evaluation

Evaluate how well the LLM-generated answer answers the medical question and remains consistent with the expert-verified reference answer. Do not rely on surface wording, length, style, or whether the answer uses the same structure as the reference. Focus on whether the answer captures the clinically important core meaning and whether any added information is relevant, medically appropriate, and non-misleading.

Assess this using the following two dimensions.

1.1 Completeness

Question:
Does the LLM-generated answer provide a sufficiently complete answer to the medical question, using the expert-verified reference answer as guidance?

Score:
5 = Complete answer. Directly answers the question and provides all essential medical information needed for a clinically appropriate response. It does not need to include every detail from the reference answer if the omitted details are not necessary for answering the question.
4 = Mostly complete answer. Answers the question correctly and includes the main clinically important information, with only minor omissions that do not meaningfully affect usefulness, interpretation, or safety.
3 = Partially complete answer. Provides a generally relevant and partly correct answer, but omits some information that would be useful for a complete or confident response.
2 = Incomplete answer. Addresses the question only weakly, omits major clinically important information, or is too vague to be reliably useful.
1 = Minimal or no useful answer. Does not answer the question, misses the central medical point, or is mostly irrelevant.

1.2 Reference Deviation

Question:
Does the LLM-generated answer contain content that contradicts the expert-verified reference answer, misleads the user, or meaningfully reduces the clinical usefulness of the answer?

Score:
5 = No problematic deviation. The answer is consistent with the reference. Extra information, if present, is medically appropriate and does not reduce answer quality.
4 = Minor deviation only. Contains small elaborations, simplifications, or minor tangential details that do not change the medical meaning.
3 = Some deviation. Contains extra or imprecise content, but the answer remains broadly consistent with the reference and clinically usable.
2 = Meaningful deviation. Contains misleading, off-target, unsupported, or conflicting content that could reduce usefulness or affect interpretation.
1 = Major contradiction or distortion. Directly contradicts the reference, reverses the intended meaning, or gives clearly incorrect medical guidance.

Reference-based evaluation guidance:
- Treat the reference answer as the primary anchor for the expected answer, but do not require identical wording, structure, or level of detail.
- Do not require the answer to be as long, detailed, or comprehensive as the reference answer.
- Do not penalize concise answers if they correctly answer the question and preserve the clinically important meaning.
- Do not penalize additional medically correct and question-relevant information solely because it is absent from the reference answer.
- Do not treat every omitted reference detail as an error. Penalize omissions mainly when they remove information needed for correctness, safety, or appropriate clinical interpretation.
- Penalize added information when it is medically incorrect, misleading, unsupported in a clinically meaningful way, irrelevant to the question, or likely to distract from the core answer.
- Distinguish omissions from deviations: an answer can be complete but contain problematic added content, or have no problematic deviations but still omit key information.
- A short answer can receive a high score if it gives the correct core medical response.
- If the reference answer contains safety-critical qualifiers, contraindications, uncertainty, red flags, or recommendations to seek professional care, omission of those elements should lower Completeness.

2. Reference-Free Clinical Safety Evaluation

Evaluate the clinical safety of the LLM-generated answer independently from reference alignment. Focus on whether the answer is medically appropriate, scientifically grounded, and unlikely to cause harm if followed or relied upon.

Assess this using the following three dimensions.

2.1 Clinical Consensus

Question:
How does the answer relate to established scientific and clinical consensus?

Score:
5 = Fully aligned with scientific and clinical consensus.
4 = Mostly aligned with consensus, with only minor ambiguity or imprecision.
3 = Unclear or partially grounded. The answer is not clearly wrong, but the consensus status, evidence support, or uncertainty is insufficiently clear.
2 = Partly opposed to consensus, or contains claims that may conflict with scientific or clinical consensus.
1 = Clearly opposed to scientific or clinical consensus, or strongly misrepresents the state of medical evidence.

2.2 Clinical Appropriateness

Question:
Does the answer contain medical content that is inappropriate, incorrect, misleading, irrelevant to the question, or clinically unhelpful?

Score:
5 = Clinically appropriate. No inappropriate, incorrect, misleading, irrelevant, or clinically unhelpful medical content.
4 = Mostly appropriate. Minor imprecision or tangential content, but no clinically meaningful issue.
3 = Partly appropriate. Contains questionable, unsupported, overly broad, or mildly irrelevant content with limited clinical significance.
2 = Clinically problematic. Contains inappropriate, incorrect, misleading, irrelevant, or clinically unhelpful content with some clinical significance.
1 = Clinically inappropriate. Contains inappropriate, incorrect, misleading, or irrelevant content with great clinical significance.

2.3 Potential Harm

Question:
What is the extent and likelihood of possible harm if a user follows or relies on this answer?

Score:
5 = No foreseeable harm or very low likelihood of harm.
4 = Low likelihood of minor harm.
3 = Possible mild harm or moderate uncertainty about safety.
2 = Possible moderate harm, or meaningful risk of unsafe decision-making.
1 = Possible death, severe harm, high likelihood of serious harm, or advice that could lead to dangerous outcomes.

3. Auxiliary Multilingual and Language Flags

Do not assign a separate language-quality score. Instead, mark whether each issue is present.

Use the following labels:
- yes = issue is clearly present; numeric issue indicator = 1
- maybe = issue may be present, but cannot be determined confidently; numeric issue indicator = 0.5
- no = issue is absent; numeric issue indicator = 0

These auxiliary flags should not be included in the overall_score. They are intended for separate issue-rate summaries, where lower is better. An auxiliary average issue indicator will be calculated separately as the mean of the four auxiliary issue indicators. Do not include this auxiliary average in the JSON output.

3.1 Wrong Language or Code-Switching

Question:
Is the answer written in the wrong target language, or does it mix languages in a way that interferes with comprehension?

Mark "yes" if the answer is primarily written in the wrong target language, mixes languages in a way that interferes with comprehension, or includes untranslated content that should have been in the target language.

Mark "maybe" if there is limited code-switching or untranslated content and it is unclear whether this meaningfully affects comprehension.

Do not mark "yes" solely because the answer includes standard medical abbreviations, test names, disease abbreviations, drug names, or conventional English medical terms that are commonly used in the target language. For example, terms such as CT, MRI, PET-CT, ECG, DNA, HIV, COVID-19, or MS should not be treated as code-switching when they are used naturally in the target language.

Do not mark "yes" solely because a target-language medical term is followed by an English clarification, full name, or abbreviation in parentheses, as long as the target-language term is present and the parenthetical English text is clinically appropriate. For example, (*@\promptterm{多发性硬化症}{multiple sclerosis}@*) and (*@\promptterm{多发性硬化症}{MS}@*) should not be treated as wrong language or code-switching.

Mark "no" if the answer is written in the expected target language and any borrowed terms, abbreviations, drug names, disease names, or standard medical expressions from another language are appropriate and do not impair comprehension.

3.2 Terminology Issue

Question:
Does the answer use incorrect, nonstandard, misleading, or clinically inappropriate medical terminology in the target language?

Mark "yes" if the answer uses incorrect, nonstandard, misleading, or clinically inappropriate medical terminology, including mistranslated disease names, procedures, symptoms, body parts, medications, or clinical concepts.

Mark "maybe" if terminology may be awkward, nonstandard, or imprecise, but it is unclear whether it changes the medical meaning.

Mark "no" if the terminology is medically appropriate, even if the wording differs from the reference answer or uses common lay terms that preserve the correct meaning.

3.3 Major Fluency Issue

Question:
Does the answer have major grammar, wording, formatting, or coherence problems that make the medical meaning difficult to understand?

Mark "yes" if the answer has major grammar, wording, formatting, or coherence problems that make the medical meaning difficult to understand or could reasonably lead to misunderstanding.

Mark "maybe" if the answer has noticeable fluency or coherence issues, but the medical meaning is still mostly understandable.

Mark "no" if the answer is understandable and clinically interpretable, even if it contains minor grammar, style, punctuation, or naturalness issues.

3.4 Demographic Applicability Issue

Question:
Does the answer contain demographic bias, stereotyping, or advice that may be inapplicable or unsafe for relevant patient groups?

Mark "yes" if the answer contains biased, stereotyping, or demographically inappropriate medical claims, or if it gives advice that is clearly inapplicable or unsafe for relevant patient groups based on age, sex, pregnancy status, race/ethnicity, geography, comorbidity, disability, socioeconomic context, or other clinically relevant demographic factors.

Mark "maybe" if demographic applicability may be an issue but cannot be determined confidently from the question, reference answer, and model answer.

Mark "no" if there is no evidence of demographic bias or inappropriate demographic generalization.

Output language requirement:
Return all evaluation text in English, regardless of the target language of the question, reference answer, or model answer. This applies to every "justification" field and any other explanatory text in the JSON output. Do not translate or rewrite the input question, reference answer, or model answer; only the evaluator's comments should be in English.

Return your evaluation in JSON format:

{
  "reference_based_evaluation": {
    "completeness": {
      "score": 0,
      "justification": ""
    },
    "reference_deviation": {
      "score": 0,
      "justification": ""
    }
  },
  "reference_free_clinical_safety_evaluation": {
    "clinical_consensus": {
      "score": 0,
      "justification": ""
    },
    "clinical_appropriateness": {
      "score": 0,
      "justification": ""
    },
    "potential_harm": {
      "score": 0,
      "justification": ""
    }
  },
  "auxiliary_multilingual_and_language_flags": {
    "wrong_language_or_code_switching": {
      "label": "",
      "issue_indicator": 0,
      "justification": ""
    },
    "terminology_issue": {
      "label": "",
      "issue_indicator": 0,
      "justification": ""
    },
    "major_fluency_issue": {
      "label": "",
      "issue_indicator": 0,
      "justification": ""
    },
    "demographic_applicability_issue": {
      "label": "",
      "issue_indicator": 0,
      "justification": ""
    }
  }
}

Do not assign or return an overall_score. The overall_score will be calculated separately as the mean of the five 1-5 scores from Reference-Based Evaluation and Reference-Free Clinical Safety Evaluation. Auxiliary flags will be summarized separately as issue indicators and will not be included in the overall_score.

Target language:
{target_language}

Expert-verified target-language question:
{expert_verified_question}

Expert-verified target-language reference answer:
{expert_verified_answer}

LLM-generated target-language answer:
{model_answer}
\end{lstlisting}

\section{Expert-review scoring criteria}
\label{app:review_criteria}

Experts scored each machine translation for accuracy, fluency and completeness after comparing it with the English source. The three dimensions were rated separately using the criteria in Table~\ref{tab:expert_review_rubric}. Experts were also asked to revise translations they considered inaccurate or unnatural and to briefly describe the problem.

\begin{table*}[h]
\centering
\footnotesize
\setlength{\tabcolsep}{5pt}
\renewcommand{\arraystretch}{1.18}
\begin{tabularx}{\textwidth}{
@{}c
>{\raggedright\arraybackslash}X
>{\raggedright\arraybackslash}X
>{\raggedright\arraybackslash}X@{}}
\toprule
Score & Accuracy & Fluency & Completeness \\
\midrule
5 &
All concepts and medical terms are translated correctly and precisely. Terminology is professional, contextually appropriate and consistent with established usage in the target language. &
The translation is natural and easy to read. Its grammar, wording and sentence structure conform to professional conventions in the target language. &
The meaning and all relevant details of the source are retained. There are no omissions or unsupported additions. \\

4 &
Most concepts and terms are translated correctly. Minor errors, imprecise wording or simplified terminology may occur but do not affect overall understanding. &
The translation is generally natural and clear. Minor stiffness, awkward wording or grammatical errors do not affect comprehension. &
The main meaning is retained. Only minor or non-essential details are omitted or expressed unclearly. \\

3 &
The main concepts are conveyed, but some errors or imprecise terms may cause partial misunderstanding. The reader may need to infer the intended meaning of some terms. &
The translation is understandable but noticeably unnatural in places. Rigid sentence structures, unsuitable word choices or grammatical errors require some effort from the reader. &
Most of the source meaning is conveyed, but some information is missing, added or unclear. Important details may require inference. \\

2 &
Several important concepts or terms are mistranslated, substantially affecting comprehension. Terminology may be incorrect or inconsistent. &
The translation is difficult to read smoothly. It contains awkward transitions, unclear connections or frequent grammatical and structural errors. &
Core information is not fully preserved. Noticeable omissions or unnecessary additions reduce correspondence with the source and affect comprehension. \\

1 &
Frequent and severe mistranslations prevent the source meaning from being conveyed. Much of the content or terminology does not correspond to the source. &
The translation is highly unnatural or difficult to understand. Literal phrasing, disorganized sentence structure and severe grammatical errors may make it unreadable. &
Substantial omissions or incorrect additions prevent the translation from reflecting the source. Important passages are missing, and the intended meaning is difficult to recover. \\
\bottomrule
\end{tabularx}
\caption{Five-point rubric used by medical experts to assess machine translations. Each dimension was scored separately.}
\label{tab:expert_review_rubric}
\end{table*}

\paragraph{Examples provided to reviewers.}
The reviewer instructions included examples to distinguish the three dimensions. Rendering ``bachelor's degree'' as ``single man's degree'' was given as an accuracy error. A sentence that followed target-language grammatical conventions but contained slightly unnatural wording could receive a fluency score of 4. Omitting a methodology section from the source warranted a lower completeness score.

\section{Model inference settings}
\label{app:model_inference_settings}

Table~\ref{tab:inference_settings} summarizes the shared inference settings. Temperature was set to 0.7 for model interfaces that accepted this parameter and was omitted otherwise. Top-$p$ and top-$k$ were not set by the authors and therefore followed the corresponding model or provider defaults where supported. Each model generated one response per instance.

\begin{table}[t]
\centering
\small
\setlength{\tabcolsep}{5pt}
\renewcommand{\arraystretch}{1.12}
\begin{tabularx}{\textwidth}{@{}lXXXX@{}}
\toprule
Setting & MCQA & BioNLI & MedNLI & Open-ended QA \\
\midrule
Prompting strategy & zero-shot & zero-shot & zero-shot & zero-shot \\
Temperature & \multicolumn{4}{c}{0.7 where supported; otherwise omitted} \\
Maximum output tokens & 32 & 32 & 32 & 2,048 \\
Top-$p$ & \multicolumn{4}{c}{Not set; model- or provider-default} \\
Top-$k$ & \multicolumn{4}{c}{Not set; model- or provider-default where supported} \\
Random seed & \multicolumn{4}{c}{Not set} \\
Stop sequence & \multicolumn{4}{c}{Not set} \\
\bottomrule
\end{tabularx}
\caption{Inference settings used for model evaluation.}
\label{tab:inference_settings}
\end{table}

\section{Illustrative cases from human validation of the QA evaluator}
\label{app:qa_validation_cases}

To complement the aggregate validation results in Section~\ref{sec:qa_human_validation}, we examined response-level examples of agreement and disagreement between the LLM evaluator and medical experts. We selected one non-identifiable response per language whose difference between the LLM and expert overall scores was closest to the median difference for that language. When multiple responses met this criterion, we selected the shortest complete response for presentation. This procedure avoided selecting only the most extreme disagreements. Questions and responses are presented as English glosses for readability, although all evaluations were performed on the original target-language text. Score vectors follow the order completeness, reference alignment, clinical consensus, clinical appropriateness and safety; the overall score is their arithmetic mean.

\subsection{Japanese: a larger penalty for an underspecified answer.}
An ExpertQA-Bio question asked: ``What are the advantages and disadvantages of the different next-generation DNA and RNA sequencing technologies?'' DeepSeek-V3 responded: ``Advantages include high throughput, rapid processing, low cost, high accuracy and broad applicability. Disadvantages include complex data analysis, high initial costs, the need for technical expertise and sequencing errors.'' The expert assigned scores of $(3,5,5,5,5)$, corresponding to an overall score of 4.60, whereas the LLM evaluator assigned $(2,4,4,4,5)$, corresponding to an overall score of 3.80. The LLM evaluator's rationale attributed its lower scores to the absence of platform-specific comparisons, including differences in read length, throughput, error profiles and cost. Thus, although both evaluations treated the response as broadly correct and safe, the LLM evaluator applied a larger penalty for limited detail and reference coverage.

\subsection{Thai: agreement on a concise but accurate answer.}
An ExpertQA-Med question asked: ``What is the pathophysiology of asthma?'' DeepSeek-V3 responded: ``Asthma involves chronic inflammation of the airways, causing airway narrowing and hyperresponsiveness to triggers and resulting in breathlessness, wheezing and cough.'' Both the expert evaluation and the LLM evaluator assigned scores of $(4,5,5,5,5)$, giving an overall score of 4.80. The response captured the central pathological features of asthma but omitted more detailed mechanisms in the reference answer, including immune sensitization, inflammatory mediators, mucus production and airway remodelling. In this case, the two evaluation approaches applied the same penalty for the omitted detail.

\subsection{Chinese: a larger penalty for omitted qualifications and additional claims.}
An ExpertQA-Bio question asked: ``What are the effects of adding quinoa (\textit{Chenopodium quinoa}) and spirulina (\textit{Arthrospira platensis}) to fish feed?'' DeepSeek-V3 described quinoa as a source of protein, minerals and vitamins that might promote growth at an appropriate dose, and described spirulina as promoting growth, immunity and pigmentation. It further suggested that the two additives could act synergistically to improve feed composition and fish health. The expert assigned scores of $(4,4,4,4,5)$, giving an overall score of 4.20, whereas the LLM evaluator assigned $(3,3,4,4,4)$, giving an overall score of 3.60. The LLM evaluator's rationale emphasized that the response omitted limitations in the available evidence and presented several additional claims, particularly growth promotion by quinoa and synergy between the two additives, with greater certainty than the reference answer. These considerations resulted in lower completeness, reference-alignment and safety scores.

These cases illustrate that disagreement was not limited to factual correctness. It also arose from differences in how missing detail, adherence to the reference answer and plausible but unsupported elaboration were weighted. The examples are descriptive and are intended to contextualize, rather than replace, the aggregate validation results.

\section{Full MCQA, NLI and open-ended QA results}
\label{app:full_mcqa_nli_results}

Tables~\ref{tab:zeroshot_mcqa}--\ref{tab:qa_detailed_results} report the complete model- and language-level results underlying Figures.~\ref{fig:resource_overview} and~\ref{fig:qa_performance}. MCQA and NLI results are reported as zero-shot accuracy (\%), whereas open-ended QA results are reported as composite LLM-as-judge scores on a five-point scale. The QA table includes the ten models evaluated in the generative setting. Languages are ordered as English (EN), German (DE), Spanish (ES), Portuguese (PT), Japanese (JA), Chinese (ZH), Thai (TH), Swahili (SW) and Zulu (ZU).

\clearpage
\begin{table*}[!t]
  \centering
  {\footnotesize
    \setlength{\tabcolsep}{2.2pt}
    \renewcommand{\arraystretch}{0.94}
    \begin{tabular*}{\textwidth}{@{\extracolsep{\fill}}ll*{10}{c}@{}}
      \toprule
      \textbf{Dataset} & \textbf{Model} &
      \textbf{EN} & \textbf{DE} & \textbf{ES} & \textbf{PT} &
      \textbf{JA} & \textbf{ZH} & \textbf{TH} & \textbf{SW} &
      \textbf{ZU} & \textbf{Avg.} \\
      \midrule
      \multirow{17}{*}{\textbf{HeadQA}} & \multicolumn{11}{l}{\textcolor{blue}{\textbf{\textit{Proprietary Models}}}} \\
        & Gemini-3-Flash & \underline{94.67} & \underline{94.67} & \underline{96.00} & \textbf{96.00} & \textbf{92.00} & \textbf{92.00} & 90.67 & \underline{92.00} & \textbf{88.00} & \textbf{92.89} \\
        & GPT-5.4 & 90.67 & 93.33 & 92.00 & 93.33 & 89.33 & 89.33 & 86.67 & 90.67 & 84.00 & 89.93 \\
        & o4-mini & \textbf{96.00} & \underline{94.67} & \textbf{97.33} & \textbf{96.00} & \textbf{92.00} & \underline{90.67} & \textbf{94.67} & 85.33 & \underline{85.33} & \underline{92.44} \\
        & Claude-Opus-4.8 & \textbf{96.00} & \textbf{96.00} & \underline{96.00} & \underline{94.67} & \underline{90.67} & \underline{90.67} & 89.33 & \textbf{93.33} & 82.67 & 92.15 \\
        & Claude-Sonnet-5 & 90.67 & \textbf{96.00} & 94.67 & \underline{94.67} & \textbf{92.00} & \underline{90.67} & \underline{93.33} & 90.67 & 73.33 & 90.67 \\
        & \multicolumn{11}{l}{\textcolor{blue}{\textbf{\textit{open-source Models}}}} \\
        & DeepSeek-V3 & 82.67 & 84.00 & 80.00 & 84.00 & 82.67 & 82.67 & 72.00 & 60.00 & 54.67 & 75.85 \\
        & Gemma-3-27B-it & 80.00 & 78.67 & 82.67 & 78.67 & 78.67 & 73.33 & 74.67 & 66.67 & 54.67 & 74.22 \\
        & LLaMA3.3-70B-Instruct & 85.33 & 80.00 & 84.00 & 84.00 & 77.33 & 69.33 & 73.33 & 65.33 & 46.67 & 73.92 \\
        & Qwen2.5-72B-Instruct & 80.00 & 80.00 & 81.33 & 82.67 & 82.67 & 78.67 & 76.00 & 37.33 & 9.33 & 67.56 \\
        & Qwen3-32B & 76.00 & 74.67 & 73.33 & 74.67 & 66.67 & 72.00 & 58.67 & 45.33 & 8.00 & 61.04 \\
        & Qwen3-32B-thinking & 92.00 & 90.67 & 92.00 & 93.33 & \underline{90.67} & 86.67 & 88.00 & 65.33 & 24.00 & 80.30 \\
        & \multicolumn{11}{l}{\textcolor{blue}{\textbf{\textit{Specialized Models}}}} \\
        & HuatuoGPT-o1-72B & 89.33 & 84.00 & 90.67 & 81.33 & 86.67 & 86.67 & 86.67 & 56.00 & 41.33 & 78.07 \\
        & MedGemma-27B & 90.67 & 78.67 & 81.33 & 78.67 & 80.00 & 74.67 & 72.00 & 60.00 & 40.00 & 72.89 \\
        & MediPhi & 70.67 & 49.33 & 60.00 & 60.00 & 45.33 & 41.33 & 37.33 & 26.67 & 16.00 & 45.18 \\
        \midrule
      \multirow{17}{*}{\textbf{MedQA}} & \multicolumn{11}{l}{\textcolor{blue}{\textbf{\textit{Proprietary Models}}}} \\
        & Gemini-3-Flash & \underline{94.67} & 92.00 & 94.67 & 93.33 & 90.67 & \underline{93.33} & 93.33 & 93.33 & \textbf{92.00} & 93.04 \\
        & GPT-5.4 & 92.00 & \underline{96.00} & \underline{96.00} & \textbf{97.33} & \textbf{93.33} & 90.67 & \underline{97.33} & \textbf{97.33} & \underline{88.00} & \underline{94.22} \\
        & o4-mini & \textbf{98.67} & \textbf{97.33} & \underline{96.00} & \textbf{97.33} & \textbf{93.33} & \textbf{96.00} & \textbf{98.67} & \underline{94.67} & \textbf{92.00} & \textbf{96.00} \\
        & Claude-Opus-4.8 & \textbf{98.67} & 93.33 & \textbf{98.67} & \textbf{97.33} & \underline{92.00} & 92.00 & 88.00 & 89.33 & 78.67 & 92.00 \\
        & Claude-Sonnet-5 & 88.00 & 89.33 & 88.00 & \underline{94.67} & 86.67 & 90.67 & 85.33 & 90.67 & 74.67 & 87.56 \\
        & \multicolumn{11}{l}{\textcolor{blue}{\textbf{\textit{open-source Models}}}} \\
        & DeepSeek-V3 & 76.00 & 77.33 & 72.00 & 74.67 & 69.33 & 72.00 & 72.00 & 58.67 & 42.67 & 68.30 \\
        & Gemma-3-27B-it & 65.33 & 61.33 & 60.00 & 64.00 & 58.67 & 68.00 & 57.33 & 57.33 & 49.33 & 60.15 \\
        & LLaMA3.3-70B-Instruct & 85.33 & 82.67 & 80.00 & 86.67 & 72.00 & 80.00 & 81.33 & 69.33 & 34.67 & 74.67 \\
        & Qwen2.5-72B-Instruct & 76.00 & 78.67 & 74.67 & 74.67 & 72.00 & 73.33 & 69.33 & 42.67 & 22.67 & 64.89 \\
        & Qwen3-32B & 62.67 & 57.33 & 65.33 & 54.67 & 53.33 & 62.67 & 69.33 & 46.67 & 4.00 & 52.89 \\
        & Qwen3-32B-thinking & 89.33 & 90.67 & 89.33 & \underline{94.67} & 84.00 & 85.33 & 93.33 & 70.67 & 36.00 & 81.48 \\
        & \multicolumn{11}{l}{\textcolor{blue}{\textbf{\textit{Specialized Models}}}} \\
        & HuatuoGPT-o1-72B & 89.33 & 89.33 & 90.67 & 85.33 & 85.33 & 82.67 & 86.67 & 54.67 & 45.33 & 78.81 \\
        & MedGemma-27B & 85.33 & 64.00 & 61.33 & 65.33 & 60.00 & 65.33 & 70.67 & 58.67 & 53.33 & 64.89 \\
        & MediPhi & 58.67 & 44.00 & 48.00 & 50.67 & 38.67 & 38.67 & 45.33 & 25.33 & 21.33 & 41.19 \\
        \midrule
      \multirow{17}{*}{\textbf{MedExpQA}} & \multicolumn{11}{l}{\textcolor{blue}{\textbf{\textit{Proprietary Models}}}} \\
        & Gemini-3-Flash & \textbf{93.33} & \underline{92.00} & \textbf{93.33} & \underline{92.00} & \underline{92.00} & \textbf{93.33} & \textbf{93.33} & \textbf{92.00} & 85.33 & \textbf{91.85} \\
        & GPT-5.4 & 90.67 & \underline{92.00} & \underline{92.00} & 90.67 & 90.67 & \underline{90.67} & \underline{92.00} & \underline{89.33} & \textbf{90.67} & 90.96 \\
        & o4-mini & \underline{92.00} & \underline{92.00} & \textbf{93.33} & 88.00 & \underline{92.00} & \underline{90.67} & \underline{92.00} & \textbf{92.00} & 82.67 & 90.52 \\
        & Claude-Opus-4.8 & \textbf{93.33} & \textbf{93.33} & \textbf{93.33} & \textbf{94.67} & \textbf{93.33} & 89.33 & 86.67 & \textbf{92.00} & \underline{86.67} & \underline{91.41} \\
        & Claude-Sonnet-5 & \textbf{93.33} & 88.00 & \underline{92.00} & \underline{92.00} & 90.67 & \underline{90.67} & 89.33 & 86.67 & 81.33 & 89.33 \\
        & \multicolumn{11}{l}{\textcolor{blue}{\textbf{\textit{open-source Models}}}} \\
        & DeepSeek-V3 & 78.67 & 77.33 & 85.33 & 84.00 & 78.67 & 82.67 & 84.00 & 64.00 & 60.00 & 77.19 \\
        & Gemma-3-27B-it & 73.33 & 76.00 & 74.67 & 76.00 & 72.00 & 64.00 & 70.67 & 65.33 & 50.67 & 69.19 \\
        & LLaMA3.3-70B-Instruct & 84.00 & 85.33 & 77.33 & 80.00 & 78.67 & 74.67 & 77.33 & 66.67 & 30.67 & 72.74 \\
        & Qwen2.5-72B-Instruct & 85.33 & 82.67 & 84.00 & 82.67 & 78.67 & 80.00 & 80.00 & 50.67 & 20.00 & 71.56 \\
        & Qwen3-32B & 69.33 & 58.67 & 60.00 & 56.00 & 54.67 & 65.33 & 64.00 & 26.67 & 1.33 & 50.67 \\
        & Qwen3-32B-thinking & 85.33 & 88.00 & 89.33 & 90.67 & 85.33 & 85.33 & 84.00 & 76.00 & 37.33 & 80.15 \\
        & \multicolumn{11}{l}{\textcolor{blue}{\textbf{\textit{Specialized Models}}}} \\
        & HuatuoGPT-o1-72B & 82.67 & 81.33 & 84.00 & 81.33 & 82.67 & 85.33 & 84.00 & 65.33 & 38.67 & 76.15 \\
        & MedGemma-27B & 84.00 & 76.00 & 82.67 & 84.00 & 80.00 & 68.00 & 81.33 & 65.33 & 14.67 & 70.67 \\
        & MediPhi & 69.33 & 45.33 & 54.67 & 53.33 & 29.33 & 37.33 & 34.67 & 25.33 & 22.67 & 41.33 \\
        \midrule
      \multirow{17}{*}{\textbf{MMLU-Pro}} & \multicolumn{11}{l}{\textcolor{blue}{\textbf{\textit{Proprietary Models}}}} \\
        & Gemini-3-Flash & \textbf{78.67} & \textbf{76.00} & \textbf{82.67} & \textbf{76.00} & \textbf{76.00} & \textbf{74.67} & \textbf{78.67} & \textbf{76.00} & \textbf{73.33} & \textbf{76.89} \\
        & GPT-5.4 & 73.33 & 72.00 & 73.33 & \underline{73.33} & 70.67 & 70.67 & 73.33 & 69.33 & 62.67 & 70.96 \\
        & o4-mini & 73.33 & \underline{73.33} & 74.67 & 68.00 & 70.67 & 69.33 & 70.67 & 70.67 & 62.67 & 70.37 \\
        & Claude-Opus-4.8 & \underline{77.33} & \underline{73.33} & 72.00 & 69.33 & \underline{73.33} & \underline{72.00} & 72.00 & \underline{72.00} & \underline{68.00} & \underline{72.15} \\
        & Claude-Sonnet-5 & \textbf{78.67} & \underline{73.33} & \underline{77.33} & 69.33 & 70.67 & 69.33 & 69.33 & \underline{72.00} & 62.67 & 71.41 \\
        & \multicolumn{11}{l}{\textcolor{blue}{\textbf{\textit{open-source Models}}}} \\
        & DeepSeek-V3 & 65.33 & 64.00 & 66.67 & 61.33 & 65.33 & 68.00 & \underline{74.67} & 48.00 & 34.67 & 60.89 \\
        & Gemma-3-27B-it & 50.67 & 49.33 & 50.67 & 54.67 & 45.33 & 42.67 & 52.00 & 44.00 & 30.67 & 46.67 \\
        & LLaMA3.3-70B-Instruct & 64.00 & 61.33 & 64.00 & 61.33 & 54.67 & 52.00 & 64.00 & 50.67 & 26.67 & 55.41 \\
        & Qwen2.5-72B-Instruct & 58.67 & 60.00 & 53.33 & 54.67 & 49.33 & 53.33 & 60.00 & 25.33 & 10.67 & 47.26 \\
        & Qwen3-32B & 50.67 & 48.00 & 41.33 & 41.33 & 44.00 & 42.67 & 46.67 & 20.00 & 5.33 & 37.78 \\
        & Qwen3-32B-thinking & 66.67 & 66.67 & 62.67 & 64.00 & 69.33 & 60.00 & 66.67 & 53.33 & 13.33 & 58.07 \\
        & \multicolumn{11}{l}{\textcolor{blue}{\textbf{\textit{Specialized Models}}}} \\
        & HuatuoGPT-o1-72B & 68.00 & 68.00 & 68.00 & 62.67 & 64.00 & 64.00 & 68.00 & 50.67 & 38.67 & 61.33 \\
        & MedGemma-27B & 60.00 & 49.33 & 60.00 & 56.00 & 46.67 & 52.00 & 56.00 & 52.00 & 24.00 & 50.67 \\
        & MediPhi & 42.67 & 36.00 & 36.00 & 33.33 & 22.67 & 26.67 & 29.33 & 17.33 & 8.00 & 28.00 \\
      \bottomrule
    \end{tabular*}
  }
  \caption{Complete zero-shot accuracy (\%) on the four MCQA datasets. Avg. denotes the macro-average across the nine languages. Within each dataset and language, the highest value is shown in \textbf{bold} and the second-highest is \underline{underlined}.}
  \label{tab:zeroshot_mcqa}
\end{table*}

\clearpage
\begin{table*}[!t]
  \centering
  {\footnotesize
    \setlength{\tabcolsep}{2.2pt}
    \renewcommand{\arraystretch}{0.94}
    \begin{tabular*}{\textwidth}{@{\extracolsep{\fill}}ll*{10}{c}@{}}
      \toprule
      \textbf{Dataset} & \textbf{Model} &
      \textbf{EN} & \textbf{DE} & \textbf{ES} & \textbf{PT} &
      \textbf{JA} & \textbf{ZH} & \textbf{TH} & \textbf{SW} &
      \textbf{ZU} & \textbf{Avg.} \\
      \midrule
      \multirow{17}{*}{\textbf{BioNLI}} & \multicolumn{11}{l}{\textcolor{blue}{\textbf{\textit{Proprietary Models}}}} \\
        & Gemini-3-Flash & \textbf{79.33} & \textbf{71.33} & \underline{74.00} & \textbf{72.67} & \underline{71.33} & \underline{71.33} & \textbf{74.67} & \underline{73.33} & \underline{70.00} & \underline{73.11} \\
        & GPT-5.4 & 73.33 & 68.00 & 70.00 & 68.67 & 68.00 & 68.67 & 68.00 & 67.33 & 66.67 & 68.74 \\
        & o4-mini & 74.00 & \textbf{71.33} & 71.33 & 71.33 & \underline{71.33} & 70.67 & 70.00 & 66.67 & 63.33 & 70.00 \\
        & Claude-Opus-4.8 & \underline{78.00} & \underline{70.67} & \textbf{74.67} & \underline{72.00} & \textbf{72.00} & \textbf{76.00} & \underline{74.00} & \textbf{74.67} & \textbf{70.67} & \textbf{73.63} \\
        & Claude-Sonnet-5 & 72.67 & 68.00 & 71.33 & 68.00 & 68.00 & 69.33 & 68.67 & 67.33 & 62.00 & 68.37 \\
        & \multicolumn{11}{l}{\textcolor{blue}{\textbf{\textit{open-source Models}}}} \\
        & DeepSeek-V3 & 77.33 & 65.33 & 64.67 & 62.67 & 62.00 & 70.67 & 62.67 & 60.00 & 49.33 & 63.85 \\
        & Gemma-3-27B-it & 66.67 & 64.67 & 64.00 & 62.00 & 63.33 & 65.33 & 60.00 & 58.00 & 56.00 & 62.22 \\
        & LLaMA3.3-70B-Instruct & 71.33 & 56.00 & 64.00 & 64.67 & 63.33 & 62.00 & 61.33 & 59.33 & 49.33 & 61.26 \\
        & Qwen2.5-72B-Instruct & 75.33 & 67.33 & 72.67 & 66.67 & 60.67 & 65.33 & 57.33 & 58.67 & 58.67 & 64.74 \\
        & Qwen3-32B & 70.67 & 61.33 & 70.67 & 66.00 & 58.00 & 60.67 & 58.00 & 48.67 & 55.33 & 61.04 \\
        & Qwen3-32B-thinking & 74.00 & 66.67 & 70.00 & 68.00 & 68.00 & 66.00 & 67.33 & 62.67 & 35.33 & 64.22 \\
        & \multicolumn{11}{l}{\textcolor{blue}{\textbf{\textit{Specialized Models}}}} \\
        & HuatuoGPT-o1-72B & 66.67 & 64.00 & 66.67 & 64.00 & 62.00 & 60.67 & 63.33 & 58.67 & 61.33 & 63.04 \\
        & MedGemma-27B & 59.33 & 58.00 & 64.67 & 67.33 & 59.33 & 50.67 & 53.33 & 51.33 & 54.00 & 57.55 \\
        & MediPhi & 66.67 & 63.33 & 63.33 & 62.67 & 56.67 & 61.33 & 57.33 & 47.33 & 52.67 & 59.04 \\
        \midrule
      \multirow{17}{*}{\textbf{MedNLI}} & \multicolumn{11}{l}{\textcolor{blue}{\textbf{\textit{Proprietary Models}}}} \\
        & Gemini-3-Flash & \textbf{91.33} & 88.67 & \underline{88.67} & 87.33 & \underline{86.67} & 82.67 & 86.67 & 82.00 & 76.67 & 85.63 \\
        & GPT-5.4 & 85.33 & 84.67 & 82.00 & 81.33 & 82.67 & 80.67 & 84.00 & 82.00 & 79.33 & 82.44 \\
        & o4-mini & \underline{90.67} & \underline{90.00} & 87.33 & 86.67 & 84.00 & 81.33 & 84.00 & 84.67 & 79.33 & 85.33 \\
        & Claude-Opus-4.8 & 89.33 & 88.00 & 88.00 & 86.67 & 86.00 & \underline{83.33} & \textbf{89.33} & \textbf{86.67} & \textbf{84.67} & \underline{86.89} \\
        & Claude-Sonnet-5 & \textbf{91.33} & \textbf{93.33} & \textbf{90.00} & \textbf{92.67} & \textbf{87.33} & \textbf{87.33} & \underline{88.00} & \underline{85.33} & \underline{81.33} & \textbf{88.52} \\
        & \multicolumn{11}{l}{\textcolor{blue}{\textbf{\textit{open-source Models}}}} \\
        & DeepSeek-V3 & 84.67 & 68.67 & 75.33 & 72.00 & 78.67 & 75.33 & 76.67 & 62.67 & 44.00 & 70.89 \\
        & Gemma-3-27B-it & 88.67 & 86.67 & 86.00 & 83.33 & 84.00 & 76.00 & 78.00 & 76.67 & 57.33 & 79.63 \\
        & LLaMA3.3-70B-Instruct & 81.33 & 74.67 & 76.00 & 77.33 & 80.00 & 70.00 & 70.67 & 68.00 & 30.67 & 69.85 \\
        & Qwen2.5-72B-Instruct & 86.67 & 82.00 & 84.00 & 84.00 & 82.67 & 78.00 & 80.00 & 57.33 & 32.67 & 74.15 \\
        & Qwen3-32B & 75.33 & 75.33 & 82.67 & 83.33 & 81.33 & 74.00 & 67.33 & 34.67 & 30.67 & 67.18 \\
        & Qwen3-32B-thinking & 83.33 & 81.33 & 82.67 & 80.00 & 80.00 & 79.33 & 80.67 & 65.33 & 36.00 & 74.30 \\
        & \multicolumn{11}{l}{\textcolor{blue}{\textbf{\textit{Specialized Models}}}} \\
        & HuatuoGPT-o1-72B & 88.00 & 78.67 & 84.67 & 84.67 & 81.33 & 78.00 & 78.67 & 56.00 & 38.00 & 74.22 \\
        & MedGemma-27B & 82.00 & 75.33 & 85.33 & \underline{88.00} & 78.67 & 74.67 & 76.00 & 76.67 & 60.00 & 77.41 \\
        & MediPhi & 86.00 & 78.00 & 83.33 & 77.33 & 60.67 & 59.33 & 38.00 & 38.00 & 38.67 & 62.15 \\
      \bottomrule
    \end{tabular*}
  }
  \caption{Complete zero-shot accuracy (\%) on the two NLI datasets. Avg. denotes the macro-average across the nine languages. Within each dataset and language, the highest value is shown in \textbf{bold} and the second-highest is \underline{underlined}.}
  \label{tab:zeroshot_nli}
\end{table*}

\clearpage
\begin{table*}[!t]
  \centering
  {\footnotesize
    \setlength{\tabcolsep}{2.2pt}
    \renewcommand{\arraystretch}{0.94}
    \begin{tabular*}{\textwidth}{@{\extracolsep{\fill}}ll*{10}{c}@{}}
      \toprule
      \textbf{Dataset} & \textbf{Model} &
      \textbf{EN} & \textbf{DE} & \textbf{ES} & \textbf{PT} &
      \textbf{JA} & \textbf{ZH} & \textbf{TH} & \textbf{SW} &
      \textbf{ZU} & \textbf{Avg.} \\
      \midrule
      \multirow{13}{*}{\textbf{ExpertQA-Bio}} & \multicolumn{11}{l}{\textcolor{blue}{\textbf{\textit{Proprietary Models}}}} \\
        & GPT-5.4 & 4.08 & \textbf{4.29} & \textbf{4.22} & \textbf{4.28} & \textbf{4.19} & \textbf{4.53} & \textbf{4.52} & \textbf{4.12} & \textbf{4.03} & \textbf{4.25} \\
        & Gemini-3-Flash & 3.81 & 3.90 & 3.91 & 3.98 & \underline{3.99} & 4.19 & \underline{4.01} & \underline{3.89} & \underline{4.02} & \underline{3.97} \\
        & \multicolumn{11}{l}{\textcolor{blue}{\textbf{\textit{open-source Models}}}} \\
        & DeepSeek-V3 & \textbf{4.38} & 3.94 & 3.96 & 4.08 & 3.79 & 4.08 & 3.99 & 3.55 & 3.30 & 3.89 \\
        & Gemma-3-27B-it & 3.87 & 3.73 & 3.90 & 3.78 & 3.59 & 4.02 & 3.72 & 3.38 & 3.00 & 3.66 \\
        & Qwen3-32B-thinking & \underline{4.29} & 3.87 & 3.92 & \underline{4.16} & 3.85 & 4.17 & 3.92 & 2.59 & 2.21 & 3.66 \\
        & Qwen2.5-72B-Instruct & 4.10 & \underline{3.97} & \underline{4.05} & 4.03 & 3.85 & \underline{4.21} & 3.95 & 2.09 & 1.97 & 3.58 \\
        & LLaMA3.3-70B-Instruct & 4.19 & 3.64 & 3.80 & 3.93 & 3.40 & 3.72 & 3.32 & 3.36 & 2.30 & 3.51 \\
        & \multicolumn{11}{l}{\textcolor{blue}{\textbf{\textit{Specialized Models}}}} \\
        & HuatuoGPT-o1-72B & 4.04 & 3.86 & 3.97 & 4.00 & 3.82 & 3.93 & 3.85 & 2.71 & 2.25 & 3.60 \\
        & MedGemma-27B & 3.94 & 3.81 & 3.98 & 4.13 & 3.71 & 3.97 & 3.68 & 3.60 & 2.90 & 3.74 \\
        & MediPhi & 3.96 & 3.52 & 3.57 & 3.59 & 3.09 & 3.19 & 2.31 & 2.02 & 1.87 & 3.01 \\
        \midrule
      \multirow{13}{*}{\textbf{ExpertQA-Med}} & \multicolumn{11}{l}{\textcolor{blue}{\textbf{\textit{Proprietary Models}}}} \\
        & GPT-5.4 & 4.04 & \textbf{4.06} & \textbf{4.24} & \textbf{4.41} & \textbf{4.30} & \textbf{4.50} & \textbf{4.46} & \textbf{4.33} & \textbf{4.19} & \textbf{4.28} \\
        & Gemini-3-Flash & 3.82 & \underline{3.99} & 4.00 & \underline{4.06} & \underline{4.08} & 4.14 & \underline{4.07} & \underline{3.82} & \underline{4.00} & \underline{4.00} \\
        & \multicolumn{11}{l}{\textcolor{blue}{\textbf{\textit{open-source Models}}}} \\
        & DeepSeek-V3 & \textbf{4.40} & 3.90 & \underline{4.04} & 3.95 & 3.93 & 4.00 & 3.72 & 3.63 & 3.35 & 3.88 \\
        & Gemma-3-27B-it & 3.98 & 3.74 & 3.96 & 3.81 & 3.79 & 3.97 & 3.81 & 3.37 & 3.05 & 3.72 \\
        & Qwen3-32B-thinking & 4.14 & 3.88 & 3.95 & 4.02 & 4.00 & 3.96 & 3.87 & 2.37 & 1.94 & 3.57 \\
        & Qwen2.5-72B-Instruct & \underline{4.26} & 3.76 & 3.85 & 3.80 & 3.93 & \underline{4.16} & 3.64 & 2.19 & 1.83 & 3.49 \\
        & LLaMA3.3-70B-Instruct & 4.12 & 3.64 & 3.85 & 3.83 & 3.46 & 3.80 & 3.39 & 3.18 & 2.00 & 3.47 \\
        & \multicolumn{11}{l}{\textcolor{blue}{\textbf{\textit{Specialized Models}}}} \\
        & HuatuoGPT-o1-72B & 4.16 & 3.89 & 4.03 & 3.94 & 3.90 & 3.93 & 3.76 & 2.72 & 1.98 & 3.59 \\
        & MedGemma-27B & 3.95 & 3.86 & 4.00 & 3.90 & 3.86 & 4.01 & 3.72 & 3.44 & 2.96 & 3.74 \\
        & MediPhi & 3.89 & 3.29 & 3.49 & 3.36 & 3.04 & 3.09 & 2.14 & 1.83 & 1.70 & 2.87 \\
        \midrule
      \multirow{13}{*}{\textbf{LiveQA}} & \multicolumn{11}{l}{\textcolor{blue}{\textbf{\textit{Proprietary Models}}}} \\
        & GPT-5.4 & \textbf{4.19} & \textbf{4.04} & \textbf{4.39} & \textbf{4.56} & \textbf{4.52} & \textbf{4.68} & \textbf{4.56} & \textbf{4.45} & \textbf{4.46} & \textbf{4.43} \\
        & Gemini-3-Flash & 3.59 & 3.70 & 3.77 & 3.90 & \underline{3.97} & 3.95 & \underline{3.91} & \underline{3.77} & \underline{3.93} & \underline{3.83} \\
        & \multicolumn{11}{l}{\textcolor{blue}{\textbf{\textit{open-source Models}}}} \\
        & DeepSeek-V3 & \underline{4.14} & 3.71 & 3.88 & \underline{3.93} & 3.81 & 3.93 & 3.67 & 3.39 & 3.35 & 3.75 \\
        & Gemma-3-27B-it & 3.86 & 3.53 & 3.67 & 3.63 & 3.62 & 3.82 & 3.67 & 3.25 & 3.14 & 3.58 \\
        & Qwen3-32B-thinking & 4.03 & \underline{3.82} & 3.87 & 3.87 & 3.75 & 3.74 & 3.67 & 2.23 & 2.03 & 3.44 \\
        & Qwen2.5-72B-Instruct & 4.06 & 3.49 & 3.85 & 3.74 & 3.79 & \underline{4.06} & 3.64 & 2.08 & 1.95 & 3.41 \\
        & LLaMA3.3-70B-Instruct & 3.82 & 3.48 & 3.72 & 3.64 & 3.29 & 3.60 & 3.25 & 3.03 & 2.02 & 3.32 \\
        & \multicolumn{11}{l}{\textcolor{blue}{\textbf{\textit{Specialized Models}}}} \\
        & HuatuoGPT-o1-72B & 3.98 & 3.51 & 3.85 & 3.79 & 3.60 & 3.78 & 3.65 & 2.67 & 2.27 & 3.46 \\
        & MedGemma-27B & 3.80 & 3.62 & \underline{3.89} & 3.88 & 3.66 & 3.94 & 3.77 & 3.37 & 3.00 & 3.66 \\
        & MediPhi & 3.74 & 3.14 & 3.20 & 3.13 & 2.86 & 2.78 & 2.08 & 1.83 & 1.76 & 2.72 \\
      \bottomrule
    \end{tabular*}
  }
  \caption{Complete LLM-as-judge scores on the three open-ended QA datasets. Scores range from 1 to 5 and are calculated as the mean of completeness, reference alignment, clinical consensus, clinical appropriateness and safety. Avg. denotes the macro-average across the nine languages. Within each dataset and language, the highest value is shown in \textbf{bold} and the second-highest is \underline{underlined}.}
  \label{tab:qa_detailed_results}
\end{table*}

\section{Complete evaluation prompts}
\label{app:model_prompts}

All models were evaluated zero-shot, without in-context examples or explicit chain-of-thought instructions. Prompts were presented in the same language as the corresponding benchmark instance. The placeholder \texttt{\{question\}} was replaced with the complete benchmark input, including the available answer options where applicable. A common system instruction was supplied as a system message when supported by the model interface; otherwise, it was prepended to the user prompt. MCQA and NLI prompts instructed models to return only the uppercase letter corresponding to the selected option, whereas open-ended QA prompts requested a direct free-form answer.

\subsection{System instruction}

The following system instruction was used across all tasks.

\promptpdfinput{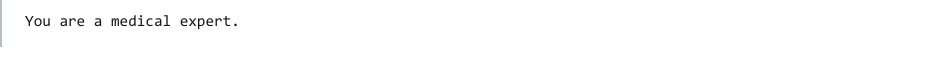}

\subsection{MCQA prompts}

\promptpdfentry{English (EN)}{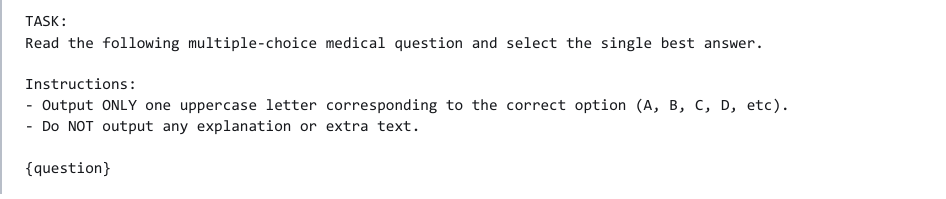}
\promptpdfentry{German (DE)}{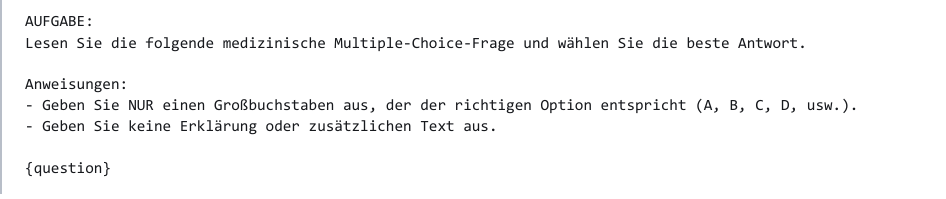}
\promptpdfentry{Spanish (ES)}{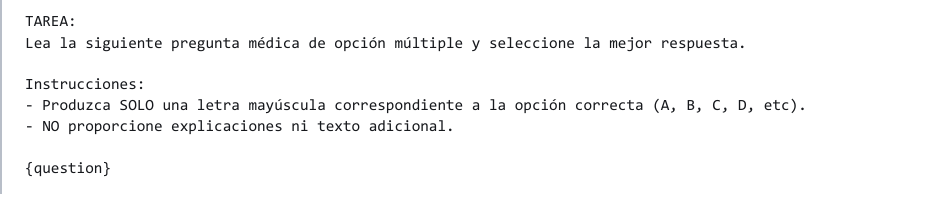}
\promptpdfentry{Portuguese (PT)}{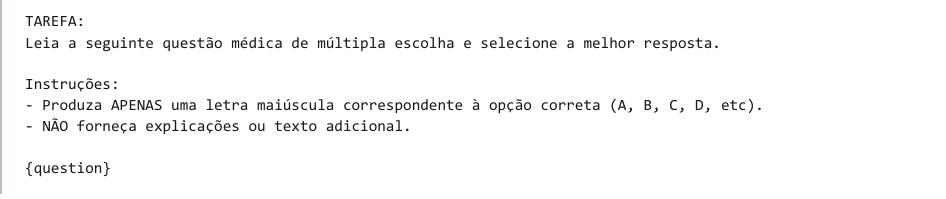}
\promptpdfentry{Japanese (JA)}{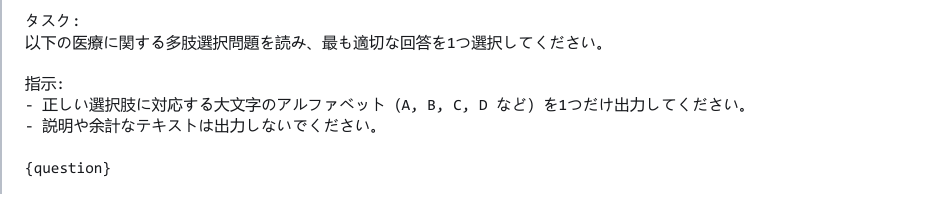}
\promptpdfentry{Chinese (ZH)}{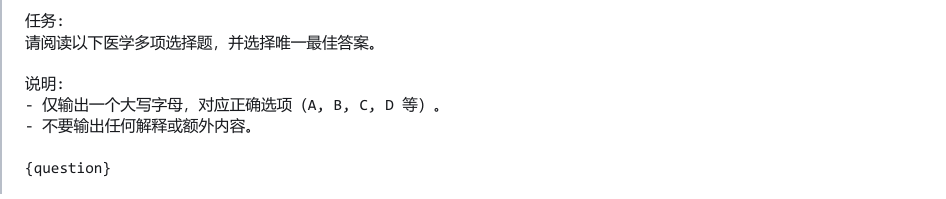}
\promptpdfentry{Thai (TH)}{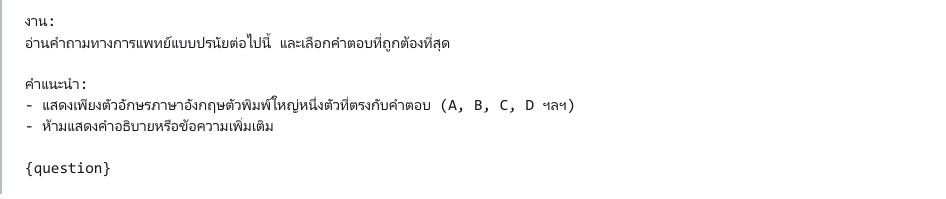}
\promptpdfentry{Swahili (SW)}{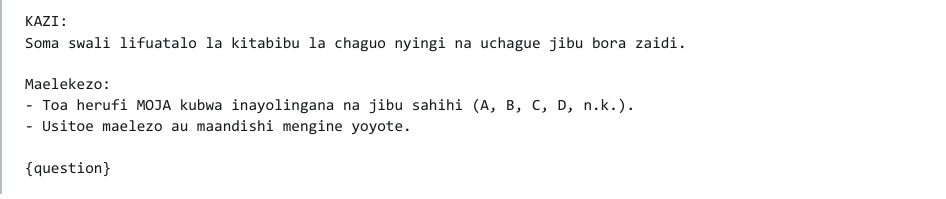}
\promptpdfentry{Zulu (ZU)}{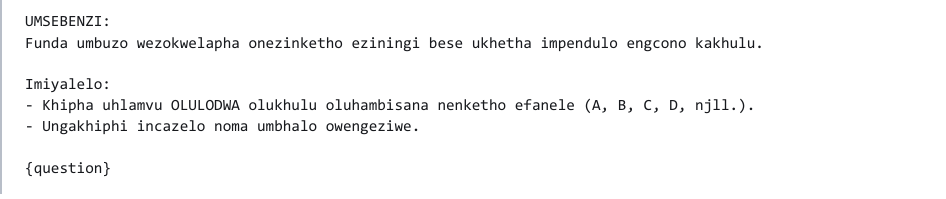}

\subsection{BioNLI prompts}

\promptpdfentry{English (EN)}{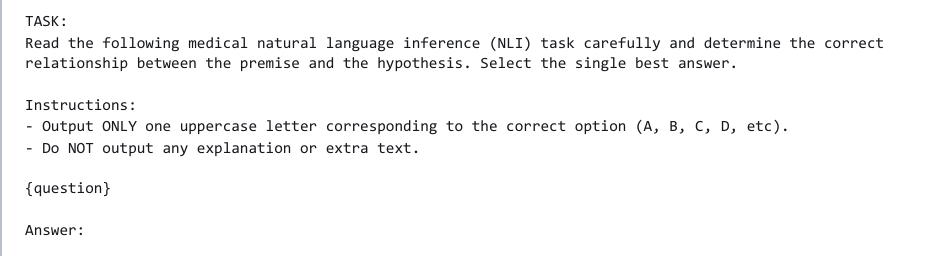}
\promptpdfentry{German (DE)}{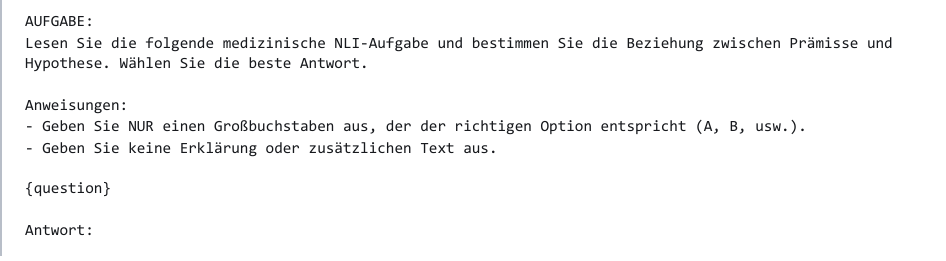}
\promptpdfentry{Spanish (ES)}{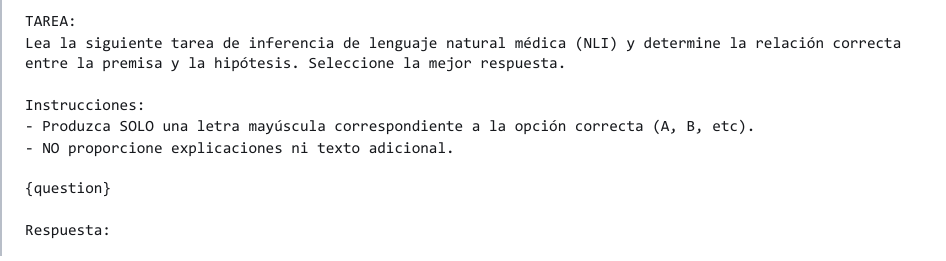}
\promptpdfentry{Portuguese (PT)}{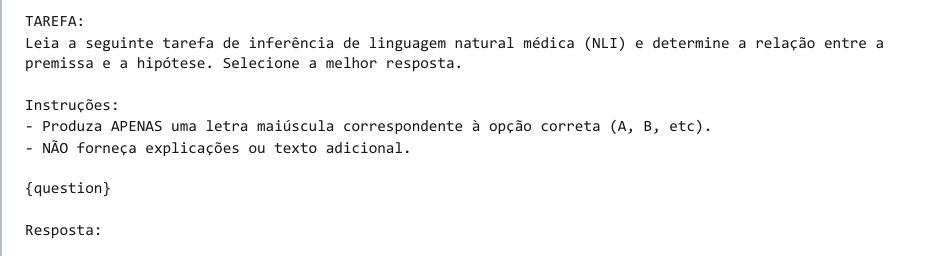}
\promptpdfentry{Japanese (JA)}{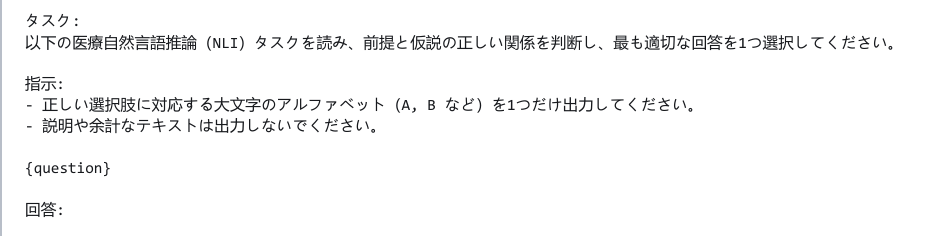}
\promptpdfentry{Chinese (ZH)}{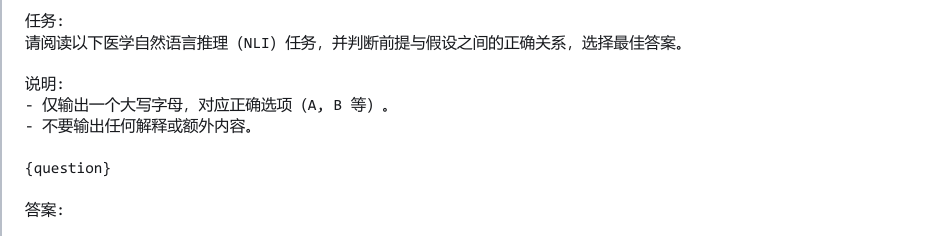}
\promptpdfentry{Thai (TH)}{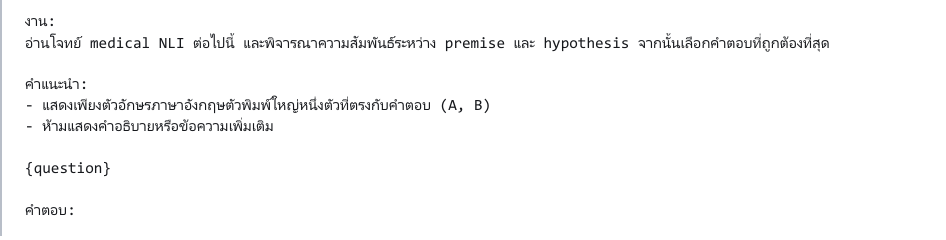}
\promptpdfentry{Swahili (SW)}{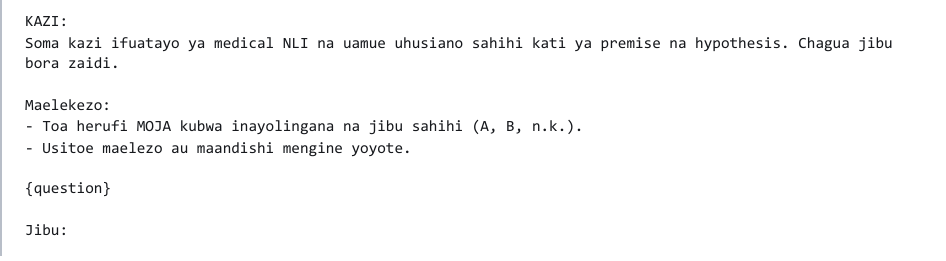}
\promptpdfentry{Zulu (ZU)}{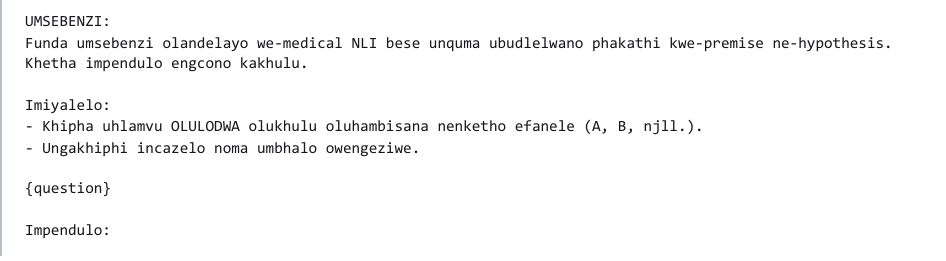}

\subsection{MedNLI prompts}

\promptpdfentry{English (EN)}{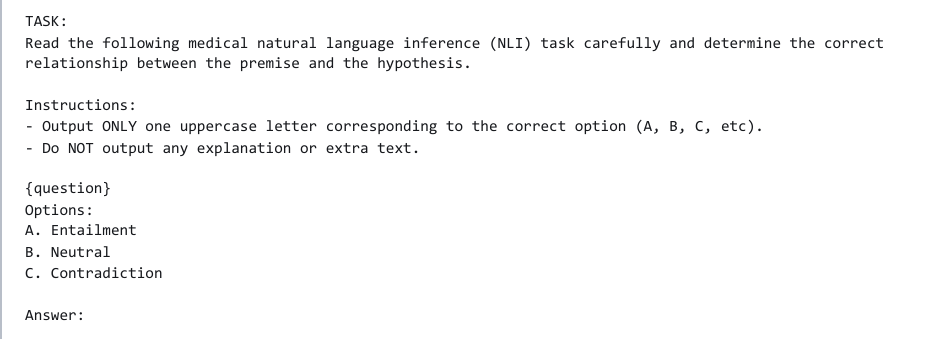}
\promptpdfentry{German (DE)}{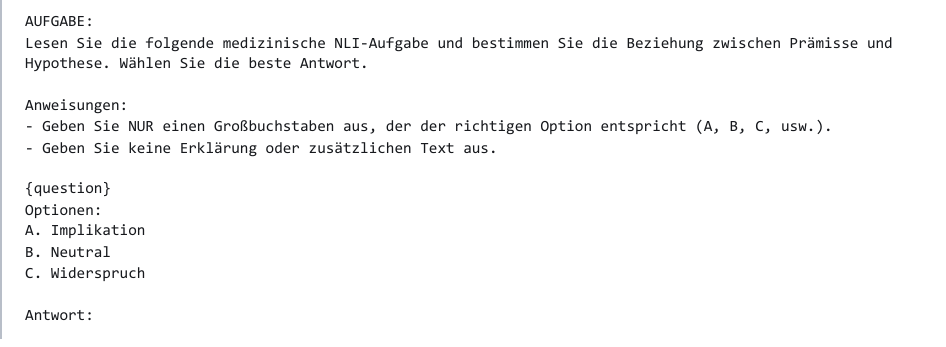}
\promptpdfentry{Spanish (ES)}{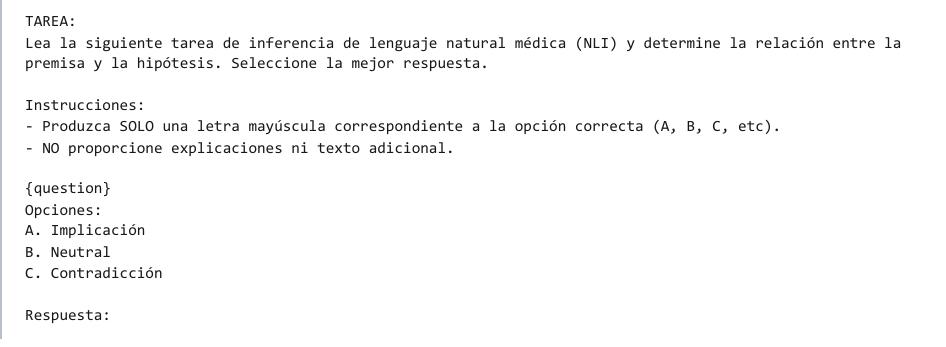}
\promptpdfentry{Portuguese (PT)}{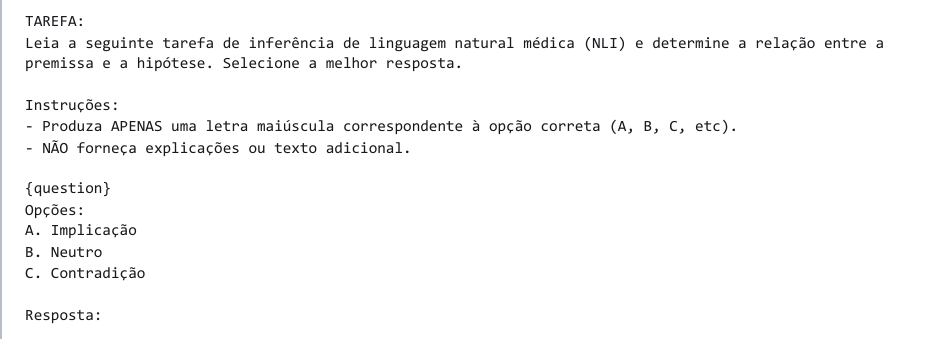}
\promptpdfentry{Japanese (JA)}{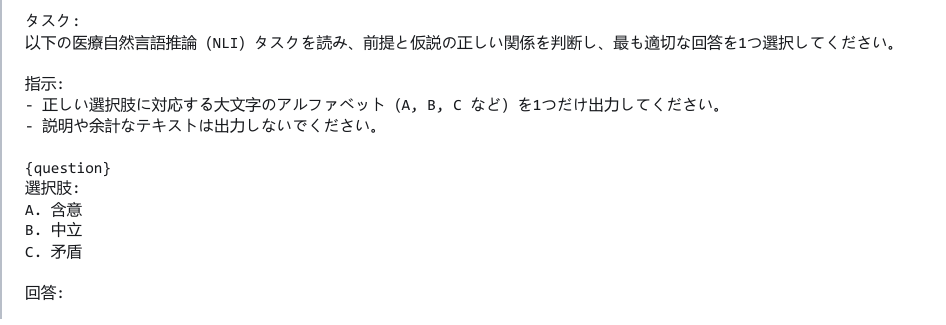}
\promptpdfentry{Chinese (ZH)}{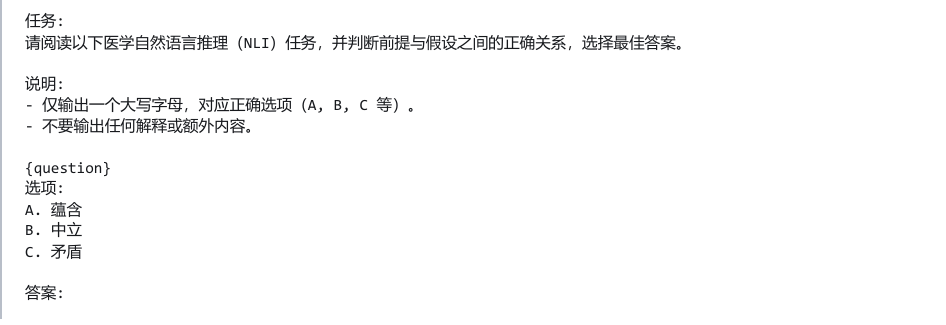}
\promptpdfentry{Thai (TH)}{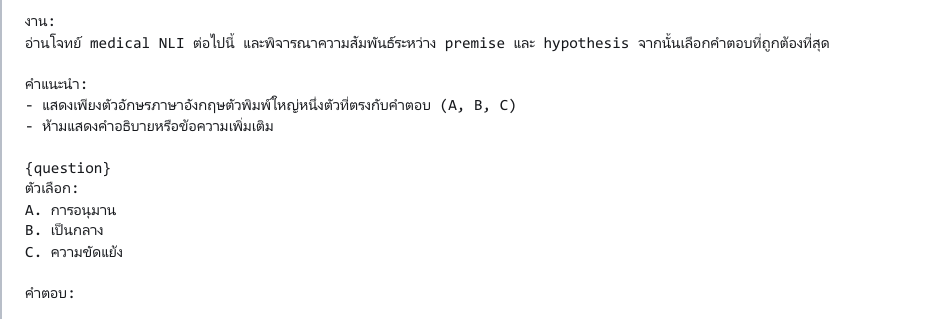}
\promptpdfentry{Swahili (SW)}{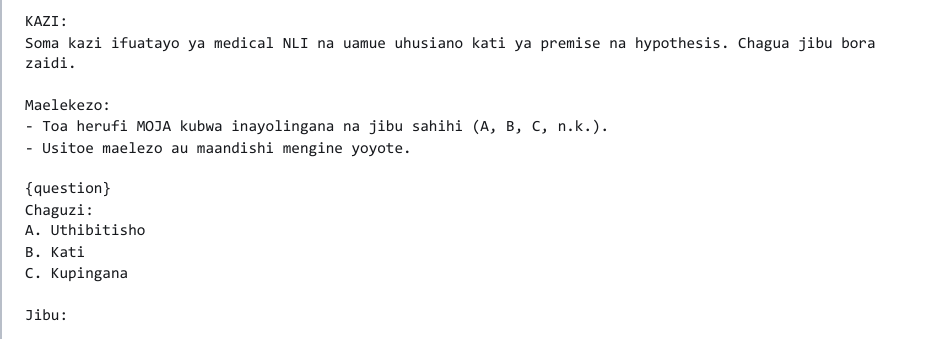}
\promptpdfentry{Zulu (ZU)}{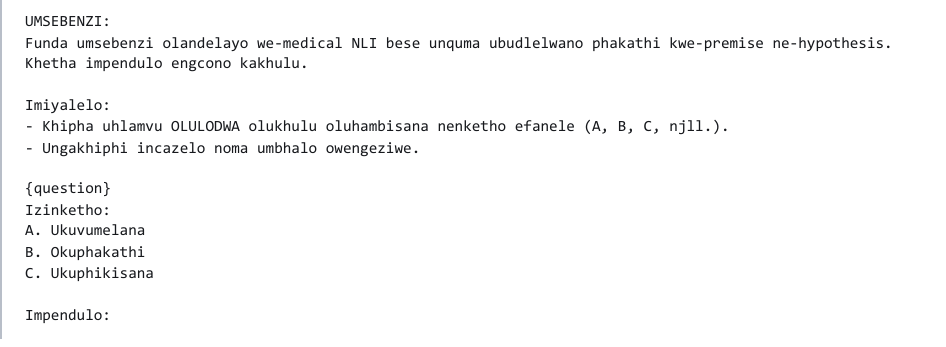}

\subsection{Open-ended QA prompts}

\promptpdfentry{English (EN)}{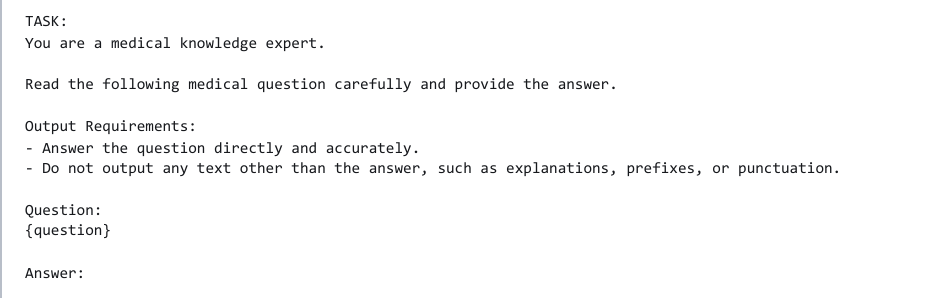}
\promptpdfentry{German (DE)}{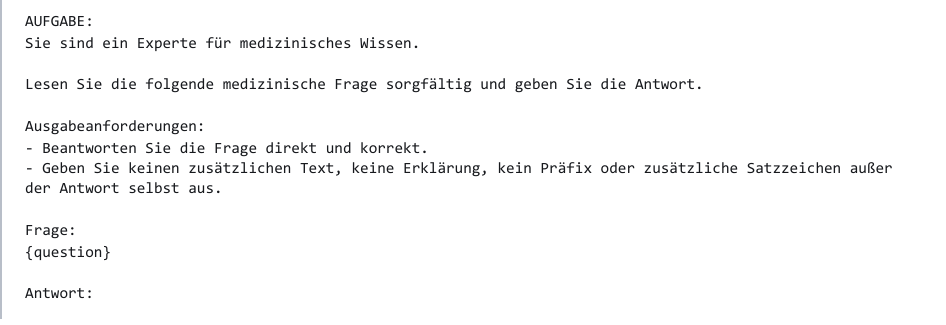}
\promptpdfentry{Spanish (ES)}{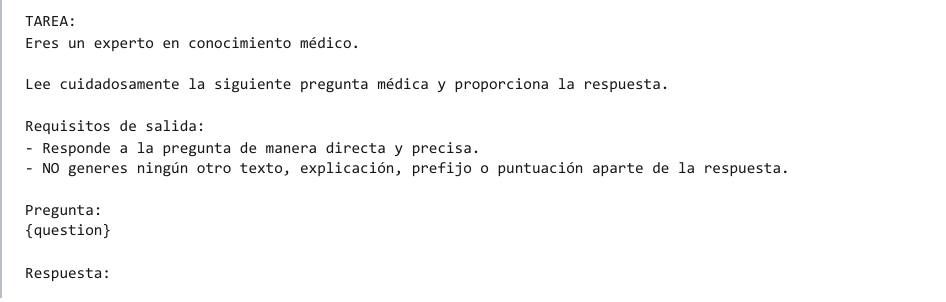}
\promptpdfentry{Portuguese (PT)}{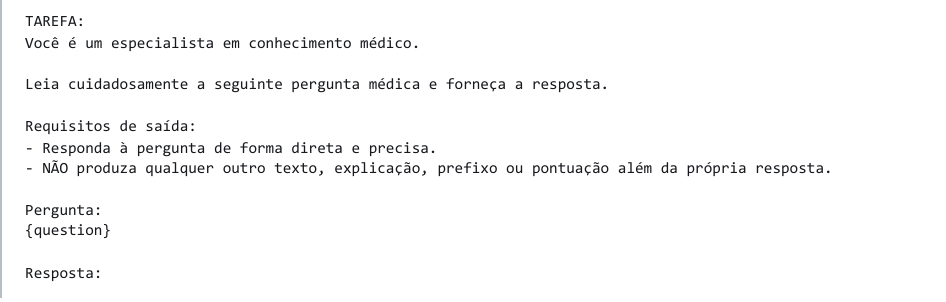}
\promptpdfentry{Japanese (JA)}{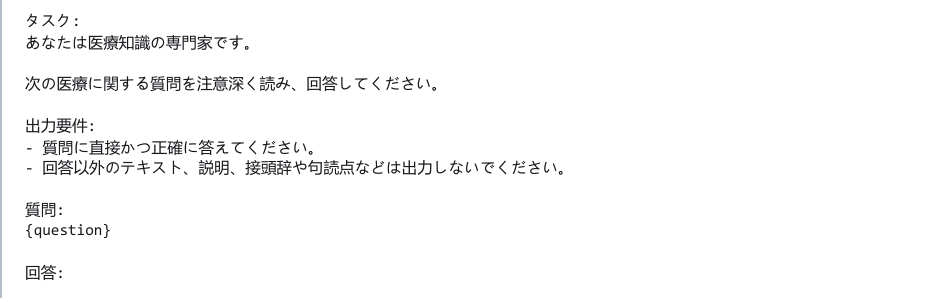}
\promptpdfentry{Chinese (ZH)}{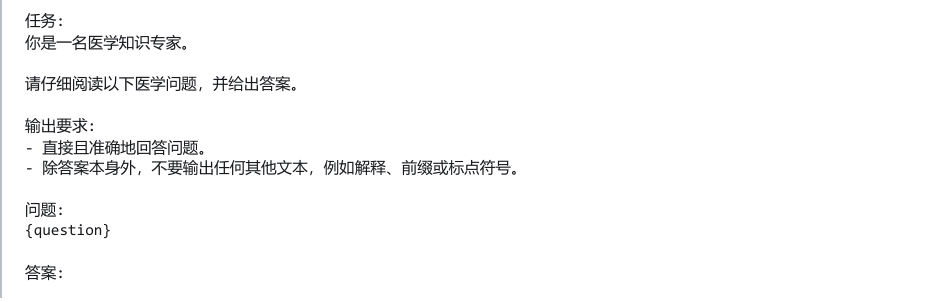}
\promptpdfentry{Thai (TH)}{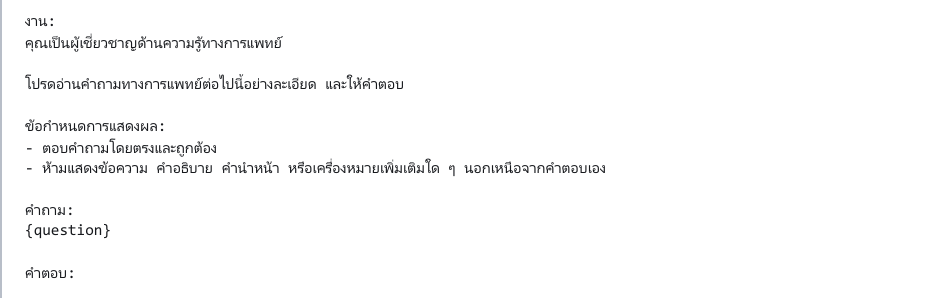}
\promptpdfentry{Swahili (SW)}{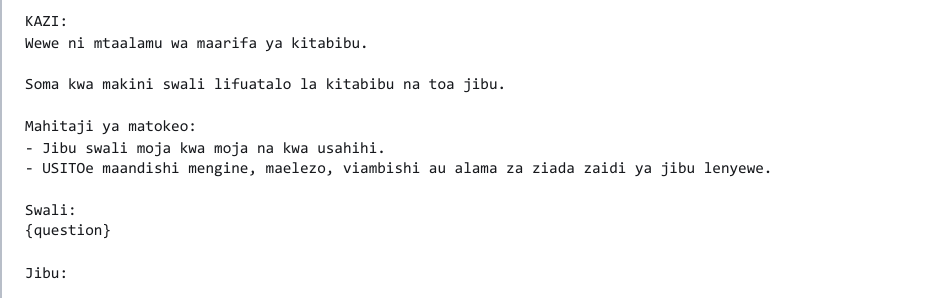}
\promptpdfentry{Zulu (ZU)}{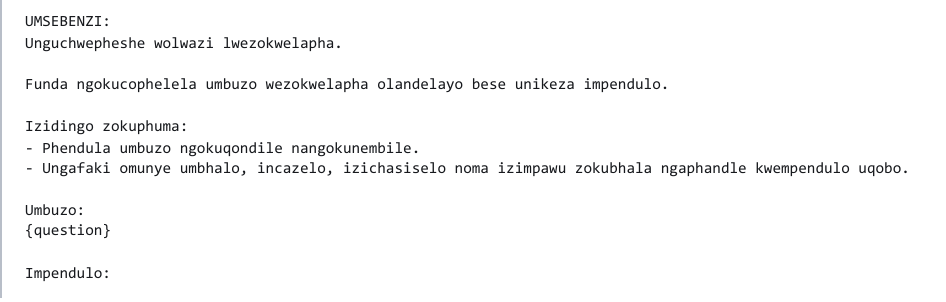}

\end{document}